\documentclass[sigconf,authorversion,natbib=true]{acmart}
\AtBeginDocument{%
  \providecommand\BibTeX{{%
    \normalfont B\kern-0.5em{\scshape i\kern-0.25em b}\kern-0.8em\TeX}}}

\if{0}
\setcopyright{acmlicensed}
\copyrightyear{2024}
\acmYear{2024}
\acmDOI{XXXXXXX.XXXXXXX}

\acmConference[SIGIR '24]{the 47th International ACM SIGIR Conference on Research and Development in Information Retrieval}{July 14--18,
  2024}{Washington D.C., USA}
\acmISBN{978-1-4503-XXXX-X/18/06}
\fi

\newcommand{\notesr}[1]{\textcolor{purple}{\small{SR: #1}}}

\newcommand{\gue}{\textsc{G4UE}}
\definecolor{aureolin}{rgb}{0.99, 0.93, 0.0}
\colorlet{lightaureolin}{aureolin!50}
\definecolor{lightgreen}{RGB}{217,255,179}

\definecolor{blush}{rgb}{255, 99,71}

\usepackage{soul} % line-breaking highlighting
\DeclareRobustCommand{\greenhighlight}[1]{{\sethlcolor{lightgreen}\hl{#1}}}
\DeclareRobustCommand{\yellowhighlight}[1]{{\sethlcolor{lightaureolin}\hl{#1}}}

\copyrightyear{2024}
\acmYear{2024}
\setcopyright{rightsretained}
\acmConference[SIGIR '24]{Proceedings of the 47th International ACM SIGIR Conference on Research and Development in Information Retrieval}{July      14--18, 2024}{Washington, DC, USA}
\acmBooktitle{Proceedings of the 47th International ACM SIGIR Conference on Research and Development in Information Retrieval (SIGIR '24), July      14--18, 2024, Washington, DC, USA}
\acmDOI{10.1145/3626772.3657882}
\acmISBN{979-8-4007-0431-4/24/07}

\begin{document}


%\title{Error-type annotation of GPT-4 responses to USMLE questions}%
\title{Beyond Accuracy: Investigating Error Types in GPT-4 Responses to USMLE Questions}


\author{Soumyadeep Roy}
\affiliation{%
  \institution{IIT Kharagpur}
  \city{Kharagpur}
  \country{India}}
%\email{soumyadeep.roy9@iitkgp.ac.in}

\author{Aparup Khatua}
\affiliation{%
  \institution{University of Michigan}
  \city{Ann Arbor}
  \country{United States}}

\author{Fatemeh Ghoochani}
\affiliation{%
  \institution{L3S Research Center}
  \city{Hannover}
  \country{Germany}}

\author{Uwe Hadler}
\affiliation{%
  \institution{L3S Research Center}
  \city{Hannover}
  \country{Germany}}

\author{Wolfgang Nejdl}
\affiliation{%
  \institution{L3S Research Center}
  \city{Hannover}
  \country{Germany}}

\author{Niloy Ganguly}
\affiliation{%
  \institution{IIT Kharagpur}
  \city{Kharagpur}
  \country{India}}

\renewcommand{\shortauthors}{S. Roy et al.}


\begin{abstract}
GPT-4 demonstrates high accuracy in medical QA tasks, leading with an accuracy of 86.70\%, followed by Med-PaLM 2 at 86.50\%. However, around 14\% of errors remain. Additionally, current works use GPT-4 to only predict the correct option without providing any explanation and thus do not provide any insight into the thinking process and reasoning used by GPT-4 or other LLMs. Therefore, we introduce a new domain-specific error taxonomy derived from collaboration with medical students. Our \textit{GPT-4 USMLE Error} (\gue) dataset comprises $4153$ GPT-4 correct responses and $919$ incorrect responses to the United States Medical Licensing Examination (USMLE) respectively. These responses are quite long ($258$ words on average), containing detailed explanations from GPT-4 justifying the selected option. We then launch a large-scale annotation study using the Potato annotation platform and recruit $44$ medical experts through Prolific, a well-known crowdsourcing platform. We annotated $300$ out of these $919$ incorrect data points at a granular level for different classes and created a multi-label span to identify the reasons behind the error. In our annotated dataset, a substantial portion of GPT-4's incorrect responses is categorized as a ``Reasonable response by GPT-4," by annotators. This sheds light on the challenge of discerning explanations that may lead to incorrect options, even among trained medical professionals. We also provide medical concepts and medical semantic predications extracted using the SemRep tool for every data point. We believe that it will aid in evaluating the ability of LLMs to answer complex medical questions. We make the resources available at \textbf{\url{https://github.com/roysoumya/usmle-gpt4-error-taxonomy}}.
\end{abstract}


\begin{CCSXML}
<ccs2012>
   <concept>
       <concept_id>10010405.10010444.10010446</concept_id>
       <concept_desc>Applied computing~Consumer health</concept_desc>
       <concept_significance>300</concept_significance>
       </concept>
   <concept>
       <concept_id>10010147.10010178.10010179.10010182</concept_id>
       <concept_desc>Computing methodologies~Natural language generation</concept_desc>
       <concept_significance>500</concept_significance>
       </concept>
   <concept>
       <concept_id>10010147.10010178.10010179.10010186</concept_id>
       <concept_desc>Computing methodologies~Language resources</concept_desc>
       <concept_significance>100</concept_significance>
       </concept>
 </ccs2012>
\end{CCSXML}

\ccsdesc[300]{Applied computing~Consumer health}
\ccsdesc[500]{Computing methodologies~Natural language generation}
\ccsdesc[100]{Computing methodologies~Language resources}


\keywords{USMLE Error Taxonomy, Medical QA, GPT-4, Multi-label Dataset}

\if{0}
\begin{teaserfigure}
  \includegraphics[width=\textwidth]{sampleteaser}
  \caption{Seattle Mariners at Spring Training, 2010.}
  \Description{Enjoying the baseball game from the third-base
  seats. Ichiro Suzuki preparing to bat.}
  \label{fig:teaser}
\end{teaserfigure}

\received{20 February 2007}
\received[revised]{12 March 2009}
\received[accepted]{5 June 2009}
\fi
\maketitle

\section{Introduction}\label{sec:intro}
Recent Large Language Models (LLMs) such as Med-PaLM 2~\cite{singhal2023medpalm2}, GPT-3.5, and GPT-4~\cite{openai2023gpt4} achieve promising performance in medical language processing applications such as medical fact-checking~\cite{wadden2020fact}, medical summarization~\cite{xie2023survey, xie2022pre}, radiology report generation~\cite{johnson2019mimic}, and medical training tool in learning health systems~\cite{chenarjun2023}. Especially GPT-4 demonstrates remarkable improvement compared to its predecessors~\cite{nori2023capabilities}. However, this enhancement has given rise to hallucination~\cite{huang2023survey} in text generation. 
 To quantify this problem, recent studies have investigated errors generated by LLMs in various domains, e.g. in answering questions from software engineering~\cite{kabir2023answers, chern2023factool} or the medical~\cite{wellbery2011flaws,dash2023evaluation,liévin2023large} domain. 
 
Here, we aim to develop an evaluation resource for the task of answering complex medical questions. We focus on the popular medical board exam dataset, MedQA-USMLE~\cite{jin2021disease} in this work. The USMLE is a multiple-choice-based examination for medical professionals seeking a license in the United States. Recent LLMs such as Med-PaLM 2 and GPT-4 achieve passing performance in terms of accuracy with $86.5\%$ and $86.7\%$ respectively on the USMLE-MedQA~\cite{jin2021disease} dataset. Simply evaluating accuracy on large QA datasets like USMLE is not enough to understand errors, as no insights into the types of errors can be gained and the types of errors as well as their sources can vary across domains~\cite{liévin2023large, chern2023factool, kabir2023answers}. 

 \begin{figure}[t]
    \centering
    \includegraphics[scale=0.35]{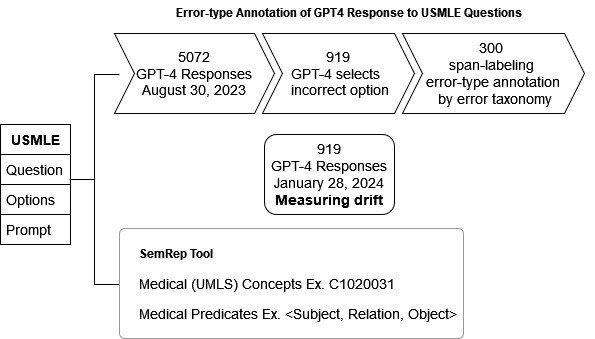}
    \caption{Resources overview of the proposed GPT-4 USMLE Error (G4UE) Dataset}
    \label{fig:resource-contrib}
\end{figure}

To delve deeper into these errors, we request longer responses from GPT-4 by prompting it to explain why it chose a particular option and why it rejected the other four options. This approach helps us understand the reasoning behind the model's decision-making process more comprehensively. We obtain responses from GPT-4 to {\bf 5072 USMLE questions}, taken from the training set of the USMLE-MedQA dataset~\cite{jin2021disease}. The responses are significantly longer (averaging 260 words) compared to prior works that only focus on predicting the correct option, then measuring accuracy \cite{chen2023chatgpt}. Additionally, the GPT-4 response contains an explanation of why it does not select the other options. This may help us in gaining precise insights and a rigorous understanding of the model's workings. To the best of our knowledge, this is the first work that explores this essential research direction. Figure~\ref{fig:resource-contrib} describes the collection of resources developed in this study for the same purpose.

We observe that GPT-4 selects the incorrect option in 919 out of 5072 cases ($18.12\%$); this subset will be the focus of the study. We propose {\bf an error taxonomy comprising seven error and two non-error categories}, with the active involvement of medical professionals in the design process. We observe that GPT-4 mostly makes reasoning mistakes leading to selection of an incorrect option. The same behavior was observed in the case of GPT-3.5 on the same MedQA-USMLE dataset where incorrect reasoning errors were the most frequent~\cite{liévin2023large}. However, our proposed error taxonomy is much more fine-grained as we introduce three new reasoning error sub-classes (where \textit{Sticking with the wrong diagnosis} and \textit{Incorrect or vague conclusion} are the most frequent) among other more detailed error classes. 

We develop the annotation setup using the open-source Potato~\cite{pei-etal-2022-potato} annotation platform and then recruit $44$ medical annotators from the Prolific platform. We maintain strict criteria for choosing medical annotators and maintain proper attention checks to remove insincere annotations. We randomly select 300 of the 919 incorrect responses and meticulously annotate them at a granular level, categorizing them into the seven error types. Additionally, we identify the specific spans responsible for the errors, presenting a multi-label span dataset. This dataset, along with the introduction of an error-type classification task, will be made publicly available, contributing to advancements in understanding and addressing inaccuracies in language models in the medical domain.

Generating this resource in an academic setting proved to be quite costly, resulting in a total expenditure of approximately GBP $1700$ throughout the process. The pay-as-you-go GPT-4 subscription model calculates its usage based on the number of input and output (generated) tokens and leads to a cost of GBP $112$ in our case. This is because our input (USMLE question, options, and prompt) is quite long (median length of 125 words) whereas the output of GPT-4 is almost double the length ($256$ words). The highest cost of GBP $1467$ was incurred during the annotation process involving $44$ medical experts to spend sufficient time to understand the annotation guidelines as well as annotate as a span-labeling task. 

GPT-4 accuracy on the USMLE dataset dropped from $86.6\%$ to $82.1\%$ on the dates of March 2023 and June 2023 respectively~\cite{chen2023chatgpt}; this behavior is formally known as \textit{performance drift}. Therefore, to verify the robustness of the proposed error categories based on GPT-4 responses on August 31, 2023, we again ask GPT-4 to respond to the 919 USMLE questions on January 28, 2024, which were previously answered incorrectly. We observe that GPT-4 still makes mistakes in $76.7\%$ of the 919 cases, highlighting that there is a lot of scope for improvement in answering complex medical questions. Additionally, this forms a useful resource for studying the performance drift of GPT-4 in complex medical questions. 
As an additional resource, we also provide standardized medical concepts and semantic predications, extracted using SemRep~\cite{Kilicoglu2020, rindflesch2003interaction}. We make the dataset and associated resources publicly available at \url{https://github.com/roysoumya/usmle-gpt4-error-taxonomy}. 
\section{Related work}
Here, we discuss recent works on LLMs for medical applications, LLM evaluation of USMLE questions, and comparison with prior error analysis works.

\noindent \textbf{Existing error analysis works in non-medical domains.} An analysis of 517 Stack Overflow questions found that 52\% of ChatGPT's responses contained inaccuracies, with 77\% being verbose. Errors include Conceptual (54\%), Factual (36\%), Code (28\%), and Terminology (12\%) issues, highlighting the need for meticulous error correction and user awareness~\cite{kabir2023answers}. %Specifically, code errors stemmed from wrong logic (48\%) and incorrect API usage (39\%), while inconsistency with human answers was notable, particularly in conceptual (67\%) and code (55\%) aspects, indicating a divergence in format and logic between ChatGPT and human-generated code \cite{kabir2023answers}.
The \textit{FacTool} framework, developed by \citet{chern2023factool} for detecting factual errors in texts generated by LLMs, identifies three specific types of errors, particularly in knowledge-based question-answering (QA). These errors include \textit{Reasoning error, Conflicting evidence, and Ambiguity} in the claim. 
 A study evaluates how well LLMs perform on the Graduate Management Admission Test (GMAT), a crucial exam for graduate management programs. By classifying errors, researchers identify specific mathematical areas where LLMs struggle, providing insights for targeted improvements. 
 Hence, it's crucial to categorize errors for better model understanding, especially when dealing with different domains. However, generic error categorization may not suffice in addressing all nuances. Therefore, we've developed more precise error categorizations tailored for the medical domain, specifically for a task like option-based question answering. This domain-specific approach helps us gain a deeper understanding of the model's capabilities and limitations, enabling us to leverage this insight for continued development and improvement.

\noindent \textbf{LLM evaluation with medical question-answering (QA).}
Recent works~\cite{wu2023pmcllama,Singhal2023} on evaluating LLMs in medical domain rely on popular QA datasets such as MedQA-USMLE~\cite{jin2021disease}, PubMedQA~\cite{jin-etal-2019-pubmedqa}, MedMCQA~\cite{pmlr-v174-pal22a} and MMLU-Medical~\cite{hendrycks2021measuring}. We focus on the MedQA-USMLE dataset for this study. The performance of state-of-the-art models on MedQA-USMLE~\cite{jin2021disease} are as follows: Galactica~\cite{taylor2022galactica} - $44.60\%$, PMC-LLaMA~\cite{wu2023pmcllama} - $44.70\%$, GatorTronGPT~\cite{Peng2023gatortron} - $42.90\%$, DoctorGLM~\cite{xiong2023doctorglm} - $67.60\%$, MedAlpaca~\cite{han2023medalpaca} - $60.20\%$, Codex~\cite{chen2021codex} - $60.20\%$, Med-PaLM~\cite{Singhal2023} - $67.60\%$, Med-PaLM 2~\cite{singhal2023medpalm2} - $86.50\%$, and GPT-4~\cite{openai2023gpt4} - $86.7\%$. 
In this work, we focus on GPT-4 as it is the best-performing model, which is closed-source and is costly to reproduce the results.

\noindent \textbf{LLM evaluation on medical tasks apart from QA.} GPT-4 provided correct diagnoses with the final diagnosis included in its differential in 64\% of cases, thereby demonstrating its potential in clinical decision support \cite{kanjee2023}. Previous research has shown the utility of LLMs in aiding generalist doctors with Differential Diagnosis (DDx) creation. However, further investigation is needed to assess their suitability for clinical settings, particularly in distinguishing specific autoimmune disorders like DIRA and FMF \cite{mcduff2023accurate}. 

\section{Research Background}

\subsection{USMLE Background and Question Format}
\label{sec:usmle-background}
The USMLE~\cite{usmlewebsite} is a three-step examination for medical licensure in the United States. It assesses a physician's ability to apply medical knowledge, concepts, and principles to patient care, ensuring that medical professionals meet the standards necessary to practice medicine in the U.S. It comprises three levels - (i) \textit{Step 1}: This is usually taken after the second year of medical school and is designed to test how well an examinee applies basic, integral science concepts to clinical scenarios, (ii) \textit{Step 2}: This is a more advanced level where the examinee's understanding of clinical science is considered essential for the provision of patient care under supervision, (iii) \textit{Step 3}: This is the most advanced level of USMLE where the examinee's medical knowledge and understanding of the basic mechanisms of disease pathogenesis is evaluated, along with treatment knowledge essential for the unsupervised practice of medicine, with the emphasis on patient management in ambulatory settings\footnote{\url{https://www.kaptest.com/study/usmle/all-about-the-usmle-step-1/}}. In this work, we also have \textit{Step} information, with two categories of USMLE questions - (i) Step 1, and (ii) Steps 2 and 3 combined.

\begin{table}[t]
%\scriptsize
    \centering
    \scalebox{0.8}{
    \begin{tabular}{|p{10cm}|} \hline 
        \textbf{USMLE question:} A 29-year-old woman, gravida 1, para 0 at 11 weeks' gestation comes to the physician because of a 2-day history of left lower extremity pain and swelling. Her temperature is 37.9°C (100.2°F). Physical examination shows a tender, palpable cord on the lateral aspect of the left lower leg. The overlying skin is erythematous and indurated. Duplex ultrasound shows vascular wall thickening and subcutaneous edema. Which of the following is the most likely diagnosis? 
        
        \textbf{Options:} {\yellowhighlight{`A'. `Deep vein thrombosis'}, `B'. `Erysipelas', `C'. `Erythema nodosum', `D'. `Varicose vein', \greenhighlight{`E'. `Superficial thrombophlebitis'}} \\
        
        \textit{Please explain your answer and why the other options are wrong} \\ \hline \hline

        \textbf{Correct Answer:} E. Superficial thrombophlebitis

        \textbf{GPT-4 Predicted Answer:} A. Deep vein thrombosis

        \textbf{ChatGPT Predicted Answer:} A. Deep vein thrombosis
        
        \\ \hline   
    \end{tabular}}
    \caption{An example of a multiple-choice question of USMLE where both GPT-4 and ChatGPT (accessed on January 28, 2024) select the incorrect option.}
    
    \label{tab:gpt4-prompt-sample}
\end{table}

\subsection{Error Taxonomies in Medical Domain}\label{sec:error-taxonomies}
\citet{wellbery2011flaws} is the closest work from the medical diagnostics domain where it provides an error taxonomy based on diagnostic errors --- \textit{Anchoring, Availability, Confirmation, Framing,} and \textit{Premature closure}. These diagnostic errors may seriously impact the decision-making ability of physicians by anchoring them to a particular diagnosis. The chain of arriving at a wrong diagnosis may involve poor intermediate reasoning steps such as relying on easily accessible information, favoring findings that confirm a diagnosis, assembling information to support a particular diagnosis, or prematurely closing the diagnostic process. 

Except for \citet{wellbery2011flaws}, all the remaining works~\cite{liévin2023large,dash2023evaluation,adams2023meta,xie2023faithful,moramarco2022human} on developing error taxonomies for answering complex medical questions are developed quite recently in the context of LLMs like GPT-3.5 and GPT-4. \citet{liévin2023large} investigated various chain-of-thought prompting techniques on GPT-3.5, and observed that most of the errors are due to the \textit{Incorrect reasoning step} (86\% of cases), followed by \textit{Incorrect or insufficient knowledge} (74\%), and a \textit{deficiency in knowledge or incorrect reading comprehension} (50\%).  In this work, we also observe the same trend where reasoning-based errors are the most frequent, with the most frequent sub-class being `sticking with the wrong diagnosis', followed by `incorrect or vague conclusion'. 

\citet{dash2023evaluation} investigate the extent of potential harm present in the output of GPT-3.5 and GPT-4. They divided the LLM responses into three distinct groups based on \textit{likelihood of patient harm} and \textit{concordance}, and then used 12 physicians to evaluate the LLM responses. Although GPT-4 performed better than GPT-3.5 overall, there is a notable absence of clear majority agreement in classifying the responses as either `Agree',  `Disagree' or `Unable to assess' for both GPT-3.5 and GPT-4 in their study. However, we observe that the mean and median inter-annotator agreement statistics, computed based on Jaccard similarity, are quite high, with values of $66 \pm 14\%$ and $64\%$ respectively, for our multi-label dataset (see Section~\ref{sec:final-annot} for further details).

 Errors may be classified into two error classes from a faithfulness perspective~\cite{xie2023faithful} --- (i) \textit{Intrinsic Error}: it occurs when the generated output contradicts existing knowledge, references, or data, (ii)  \textit{Extrinsic Error}: it cannot be confirmed by existing knowledge or references. In our proposed error taxonomy, the knowledge-based error classes of non-medical factual error and unsupported medical claim are based on similar principles and thus show that our proposed error taxonomy covers such faithfulness-based errors.

\section{Dataset Preparation}\label{sec:prepare-dataset}
Here, we first describe the construction of GPT-4 responses to USMLE questions in Section~\ref{sec:gpt-insights} and then describe the complete annotation setup using Potato~\cite{pei-etal-2022-potato} in Section~\ref{sec:annot-setup}. Section~\ref{sec:error-taxonomy} then describes the proposed error taxonomy of GPT-4 responses to complex medical questions and its development process, followed by the annotator details recruited through Prolific in Section~\ref{sec:annotator-details}. We describe the procedure of combining labels from multiple annotators in Section~\ref{sec:final-annot}. Section~\ref{sec:add-resources} details the associated resources that we develop for a more comprehensive evaluation of GPT-4 performance in answering complex medical board exam questions.

\subsection{GPT-4 Insights: Tailored Responses for USMLE Questions}\label{sec:gpt-insights}

We start with the MedQA paper \cite{jin2021disease} which provided a large dataset of $12723$ questions of the USMLE. We consider only the training data split in this study that consists of $10178$ data points. We run GPT-4 on all these data points by using the GPT-4 \textit{Chat Completions} API with a temperature of $0.8$. During the experiment, two primary versions of OpenAI's API, namely GPT-4 and GPT-3.5, were available. For our assessments conducted on 31st August 2023, we exclusively utilized GPT-4. To streamline the process, we interacted with these services using a straightforward prompt. We provide an example of such a prompt in Table~\ref{tab:gpt4-prompt-sample}.

 \subsubsection{Prompt selection and justification}
 We give the following prompt: \textit{`USMLE question:' } [question with answer options] \textit{`Please explain your answer and why the other options are wrong'}. We choose this prompt not only to improve performance through chain-of-thought reasoning but also to elicit the model to not only choose a single answer but also explain its reasoning in opting for or against each of the options. This allows for a more rigorous understanding of the model's decision process and, therefore, more insight into why errors are made.

\subsubsection{Analysis of GPT-4 responses} We received GPT-4 responses in $49.8\%$ cases ($5072$ out of $10178$ training data points). The GPT-4 generates responses presented as comprehensive paragraphs, each averaging $257.9 \pm 45.5$ words in length, with the median at $256$ words. The mean and median values of sentence count are $10.9 \pm 2.4$ and $11$ respectively. It contained $2771$ Step 1 and $2301$ Step 2 and 3 questions. We thus observe that the GPT-4 responses are almost double the length of USMLE questions in terms of median word count. The average and median word count of USMLE questions is $131.0 \pm 51.2$ and $125$ words respectively, while the same amounts to $8.0 \pm 3.2$ and $8$ sentences respectively.

\subsubsection{Construction of GPT-4 Response Dataset that selects the incorrect option.} Given that the GPT-4 response is a paragraph of continuous text, extracting the correct option is accomplished through meticulous scrutiny of string patterns using regular expressions as well as manually. First, we extract the sentence containing the correct option by using the following list of phrases: \textit{correct answer, best answer, best answer is, answer is, therefore, answer, answer would be, most appropriate, answer: , most likely finding, most likely answer, most appropriate treatment, most likely cause, most likely diagnosis, most likely diagnosis is, most likely diagnosis would be}. Given the extracted sentence containing the predicted answer, we use the following patterns to extract the predicted option: \textit{correct answer is \_, best answer is \_,  (choice \_, (option \_ }. However, this only covers around $80\%$ of all data points. For the remaining $20\%$, we manually extract the predicted option from the response. 

\subsubsection{Analysis of GPT-4 Responses that select the wrong option.} Notably, our analysis reveals that among $5072$ data points, $919$ instances ($18.1\%$) exist where GPT-4 selects the inaccurate option (wrong answer). These data points form the focus of this study and comprise $462$ Step 1 and $457$ Step 2 and 3 USMLE questions. We observe that GPT-4 has a higher percentage of incorrect responses ($19.9\%$ versus $16.7\%$) in the case of Steps 2 and 3, as compared to Step 1. This indicates the nature of human difficulty is also reflected in the machine-perceived notion of difficulty in answering USMLE questions. We observe that the mean and median length of this subset of GPT-4 responses is $268.2 \pm 47.0$ and 266 words respectively. These responses are much longer (median and $75$ percentile word count increases by $3.82\%$ and $4.2\%$ respectively) than the responses from the full dataset (5072). 
We observe that the mean and median length of this subset of USMLE Questions is $136.0 \pm 53.6$ and 128 words respectively. These USMLE questions are marginally longer than the questions from the complete set of 5072 data points. 

\subsection{Annotation Setup}\label{sec:annot-setup}
For error-type annotation, we randomly select $300$ data points out of $919$ data points where GPT-4 selects the incorrect option. We develop the annotation platform using Potato~\cite{pei-etal-2022-potato}, an open-source annotation software, and obtain the medical experts through the Prolific~\cite{prolificwebsite} crowd-sourcing platform.

\subsubsection{Methodology for developing annotation guidelines} 
Given the emerging and expert domain nature of annotating GPT-4 responses, we involve two recently graduated medical students in designing our annotation guidelines and also refining the error categories. After multiple iterations and updating the guidelines accordingly,  we finalize it. We next conduct a pilot study on Prolific, where we ask participants from medical backgrounds to annotate the error categories. We also ask them to give feedback on the error categories, which we then utilize to further refine our annotation setup and flag any gross errors or error categorization issues. We again utilize the same two recent medical graduates to better understand the feedback and confirm whether the issues raised are appropriately addressed from the medical expert knowledge perspective. This multi-stage guideline development makes us confident that no gross errors or gross response incorrectness exists. 

\subsubsection{Annotation Setup}
We added more study-specific questions based on the feedback from medical collaborators and the pilot study about the highly interdisciplinary nature of the annotation (medical experts evaluated AI-generated medical responses to complex USMLE questions). This is aimed at deeply inspecting the medical expertise of the annotators as well as the relevant clinical background required to evaluate responses to USMLE questions. We believe such detailed reporting on annotators (mostly missing in recent works) is essential to judge the veracity of the annotations. The questions are as follows: (i) \textit{Have you taken and passed the USMLE or a similar medical licensing examination?} (ii) \textit{Were you already familiar with the format and content of the USMLE examination before joining this study?} (iii) \textit{Have you worked with AI or natural language processing technologies in a medical context before?} (iv) \textit{Are you currently practicing medicine, or have you previously practiced in a clinical setting?}

\subsubsection{Annotation Process}
The task entails understanding the reason for the incorrect GPT-4 response and assigning the response to one or more of the seven error categories. There is also the option to choose from two non-error categories. However, once a non-error category is selected, annotators cannot opt for any other error category. The annotation guidelines document furnishes a formal definition to represent an error category and provides one or more positive examples for each category, along with an explanation. Each data point is annotated by three annotators. Each annotator is assigned to annotate 20 data points in $100$ minutes and is compensated at a rate of $13.20$ UK pounds per hour. We used the Potato annotation platform, and the question order was randomly shuffled to avoid position bias.

We design the annotation as a span-labeling task on the Potato annotation platform where the annotator is asked to select any span of text in the GPT-4 response and assign any of the error types. Multiple textual spans with different error types can be labeled in a single GPT-4 response. This results in a multi-label classification dataset. Table~\ref{tab:multi-label-example} shows an example of such a multi-label span annotation for a GPT-4 generated response. This table offers an illustration of how various labels are assigned to different spans within the generated long response. The annotator is provided with a USMLE question along with multiple-choice options, prompt, and correct answer that is given as input to GPT-4, as well as the response from GPT-4. 

\subsection{Proposed Error Taxonomy}\label{sec:error-taxonomy}
We propose seven error and two non-error types that the annotators may utilize if the identified error cannot be assigned to our seven error categories. In Table \ref{tab:pos-exam-reason-errors}, we elucidate each error category along with its respective explanations for the response generated by GPT-4. This presentation aims to offer a clearer and more detailed understanding of the various types of errors encountered in the generated text.  We assign these (non-)error types to four main groups and define them as follows: 

\subsubsection{Reasoning-based Error}
These errors refer to mistakes that arise mainly due to a flawed application of logic, including flawed or inconclusive reasoning processes as well as inconsistent or contradictory reasoning. We propose the following three error types:

\noindent \textbf{Error 1: Sticking with the wrong diagnosis.}
The model confidently states its (wrong) answer early in the response. It's reasoning when explaining this choice and the other options are sound and factual concerning the diagnosis, but the model still chooses the wrong answer it has given before (instead of correcting it based on the explanations it has given afterward).

\noindent \textbf{Error 2: Incorrect or Vague conclusion.} 
The explanations given by the model are factually correct and support the proposed answer, but the statements concluding the correctness of option(s)  are not definitive. Instead, vague terms are used (e.g. usually, most of the time) and the model prioritizes only a few symptoms, leading to the wrong conclusion. Another scenario is that the difference in reasoning in the response between the GPT-4 chosen option and the ground-truth reference answer is not clearly mentioned and sufficient explanation is not provided in this context.

\noindent \textbf{Error 3: Ignore missing information.}
The model recognizes that a critical piece of information is missing that would be needed to answer the question (e.g. a missing CT scan) and states so. Nevertheless, it continues to try and answer the question with incomplete information. Another scenario is when it does not recognize (mention) that a resource mentioned in the question is missing.

\subsubsection{Knowledge-based error}
In contrast to the previously presented error types, knowledge-based errors arise due to inaccurate factual knowledge or incomplete understanding of the context. We split such errors based on the underlying domain:

\noindent \textbf{Error 4: Non-medical factual error.}
A statement in the explanations of the answer is factually wrong (e.g. $1+1=3$). These involve non-medical facts including computational errors and terminology errors and can be reliably annotated by a non-medical expert.

\noindent \textbf{Error 5: Unsupported medical claim.}
One or multiple medical claims in the model's answer are wrong, i.e. not supported by proper evidence from textbooks or other online knowledge sources. One scenario might be that the model states a symptom from the USMLE question is not typically associated with a certain illness or diagnosis, while in reality, it would be.

\subsubsection{Reading comprehension error}
This error type is characterized by the model being unable to take account of all information that was given in the question or of all the tasks it was given. Here, we include the model making up information, but not ignoring information (if it recognizes information was missing, as stated before), and propose two error types:

\noindent \textbf{Error 6: Incorrect understanding of the task.} The model did not choose a single option as the correct answer. Instead, it either considers multiple answers to be correct (or most likely), states all options are wrong or does not decide overall. Another scenario is when the response does not explain why the other options are wrong, as it was tasked to do so through our prompt.

\noindent \textbf{Error 7: Hallucination of information.} In the model's argumentation for or against different options, it fabricates some information that was not included in the question to justify why an option is or is not correct. For example, GPT-4 introduces a new symptom that is not present in the USMLE question. 

\begin{table*}[!ht]
    \centering
    \scalebox{0.55}{
    \begin{tabular}{p{1.7cm}p{19.5cm}p{6cm}} \hline
          \textbf{Error Type}& \textbf{GPT-4 Response}&\textbf{Explanation of GPT-4 Response}\\ \hline
          \textbf{Error 1: Sticking with Wrong Diagnosis} Question ID: \textit{78}, Correct Answer: \textit{(A) Decreased total body potassium} & The boy's polyuria (as evidenced by bedwetting), polydipsia (increased water drinking), and unexplained weight loss, along with the presence of glucose and ketones in the urine, suggests new-onset type 1 diabetes mellitus (DM1). \yellowhighlight{Prolonged hyperglycemia leads to osmotic diuresis, dehydration, and eventually diabetic ketoacidosis (DKA) if not treated}. DKA is characterized by hyperglycemia (serum glucose >250 mg/dL), ketosis, and metabolic acidosis. The clinical features of DKA  include polyuria, polydipsia, nausea/vomiting, abdominal pain, Kussmaul respirations (ie, rapid, deep breathing), and lethargy to coma. The \yellowhighlight{diagnosis of DKA} is confirmed by arterial blood gas (ABG) analysis showing a low pH (acidemia) and low bicarbonate concentration due to the metabolic acidosis.  Option E, 'Serum glucose concentration > 600 mg/dL', is the correct answer.  Options A, B, D could be seen in different conditions but not in \yellowhighlight{DKA}.  Option A: Total body potassium is usually high in \yellowhighlight{DKA} as a result of the acidosis, not decreased.  Option B: Total body sodium can be low, normal, or slightly high during DKA, but not significantly increased.  Option D: Hypervolemia is not a feature of \yellowhighlight{DKA} which usually present with dehydration due to osmotic diuresis. Option C, 'Increased arterial pCO2', is incorrect because the body would be compensating for the metabolic acidosis by hyperventilating (Kussmaul respirations)  in an attempt to lower the CO2 level, not increase it. Hence the pCO2 would be expected to be low, not high. &  GPT-4 identifies factually correct disease symptoms but introduces the new (noisy) idea of prolonged hyperglycemia,  i.e., if the symptoms are allowed to persist for a long time, the patient may eventually develop diabetic ketoacidosis (DKA) in the future. However, the reasoning provided by GPT-4 is entirely (overly) centered around DKA, and ignores important parts of the USMLE questions. \\ \hline
          \textbf{Error 2: Incorrect or Vague Conclusion} Question ID: \textit{25}, Correct Answer: \textit{(B) Urinary pH: 4.7} & \greenhighlight{This patient's severe, colicky, right-sided flank pain that radiates to his right groin is characteristic of nephrolithiasis (kidney stones). His history of being treated with naproxen for swelling and pain in his right toe indicates gout,} \yellowhighlight{which often is associated with uric acid nephrolithiasis, as both conditions are due to overproduction or underexcretion of uric acid.} \greenhighlight{However, the most common type of nephrolithiasis is calcium oxalate, which can occur in patients with hypercalciuria from any cause.} \yellowhighlight{Uric acid stones are radiolucent and may not be seen on an x-ray, as in this patient. Calcium oxalate stones are more likely to be seen on x-ray, but a negative x-ray does not rule out the presence of a stone, as shown in this patient's CT scan.} \greenhighlight{Patients with uric acid stones often have acidic urine (pH < 5.5). In contrast, calcium oxalate stones are more likely to form in alkaline urine (pH > 7.0).} Option B: Urinary pH: 4.7 is incorrect because it is associated with \yellowhighlight{uric acid stones, which are less likely given this patient's history and CT findings.} Option C: Hexagon shaped crystals are associated with cystinuria, a genetic disorder that results in defective renal tubular transport of cystine, leading to recurrent nephrolithiasis. This patient has no history of recurrent stones, making this option less likely. Option D: Positive nitrites test indicates a urinary tract infection, which typically presents with dysuria, urgency, and frequency, symptoms that this patient does not have. Option E: Largely positive urinary protein can be seen in nephrotic syndrome, which typically presents with edema and hyperlipidemia, signs absent in this patient. & Here, GPT-4 (correctly) identifies kidney stones as the underlying condition of the patient, and the history indicates gout, which results from an imbalance in uric acid production. “No abnormalities are shown in the x-ray", as stated in the question, GPT-4 states that calcium oxalate stones are more likely to be seen in an x-ray as compared to uric acid stones; thus suggestive that uric acid stones are more likely. However, a CT scan of the abdomen and pelvis shows the presence of stones. Here, GPT-4 vaguely concludes without clear justification as to why calcium oxalate stones are more likely in this case. %If it had gone for uric acid stones, GPT-4 based on its own logic, would have opted for the correct option (Urinary pH: 4.7). 
          \\ \hline
          \textbf{Error 3: Ignore Missing Information} Question ID: \textit{19}, Correct Answer: (B) Bullous changes of the lung bases on chest CT & \yellowhighlight{Without the actual image of the liver biopsy and further clinical information, it is challenging to definitively answer this question. However, basing the answer on the given details and the options provided, the most likely answer would be 'E': 'Dark corneal ring on slit-lamp examination} The patient's age, his symptoms (jaundice, dark urine), his lab findings (elevated ALT and AST), and the use of periodic acid-Schiff-staining in his liver biopsy suggest liver disease, possibly due to a metabolic disorder such as Wilson's disease. Wilson's disease is a rare autosomal recessive disorder characterized by defective copper transport, leading to copper accumulation in various tissues, including the liver, brain, and cornea. The dark corneal ring (Kayser-Fleischer ring) is a classic ocular manifestation of this disease. Now, let's discuss why the other choices are unlikely: Option 'A' refers to Whipple disease, a systemic disorder caused by the bacterium Tropheryma whipplei. While it can cause malabsorption and weight loss, it is not typically associated with liver disease or the findings described. Option 'B' refers to conditions such as COPD or Alpha-1 Antitrypsin Deficiency, which would typically present with respiratory symptoms, not just liver disease. Option 'C' describes a cholangiogram seen in primary sclerosing cholangitis (PSC), a disease that typically presents with a history of inflammatory bowel disease, which is not mentioned here. Option 'D' refers to iron overload conditions like Hemochromatosis. While this condition can cause liver disease, myocardial iron deposition would be an uncommon initial presentation, and the patient’s described presentation does not align well with hemochromatosis. Remember that while the most likely answer based on the provided information is Wilson's disease, the diagnosis would typically require additional tests like serum/urinary copper and ceruloplasmin levels. Also, the liver biopsy in Wilson's disease would usually show steatosis, glycogenated nuclei, and copper deposits.& GPT-4 understands that the actual image of the liver biopsy and further clinical information are missing. However, it continues to provide an answer and justification. The relevant portion is \yellowhighlight{highlighted}. \\ \hline
           \textbf{Error 5: Unsupported Medical Claim} Question ID: \textit{17}, Correct Answer: (D) Cardiac stress test & \yellowhighlight{This patient's abdominal discomfort with exertion and postprandial discomfort may suggest "abdominal angina" due to chronic mesenteric ischemia (CMI). CMI is most frequently observed in patients with} \textcolor{red}{generalized atherosclerotic cardiovascular disease which can be surmised from this patient's history of smoking, peripheral arterial disease}, \yellowhighlight{hypertension, and type 2 diabetes mellitus. The splanchnic circulation has a rich collateral flow. Consequently, at least two of the three major splanchnic vessels (celiac trunk, superior mesenteric artery, inferior mesenteric artery) must be significantly stenotic before symptoms of CMI develop. This patient's postprandial abdominal pain is due to an inability to augment splanchnic blood flow in response to increased metabolic demand after meals. As a result, patients may eat less and lose weight. The pain after a "big meal" and his inability to gain symptomatic relief despite a significant decrease in food intake due to fear of provoking the pain support this diagnosis}. \textcolor{red}{Atherosclerosis} \yellowhighlight{in other vascular beds can cause symptoms (eg, claudication), but this patient's discomfort with walking is more likely to be due to his deconditioning and obesity than to his stage 2 peripheral arterial disease, which is typically asymptomatic. Computed tomographic angiography (CTA) of the abdomen is the best initial diagnostic test for CMI. Therefore, Answer 'B' (CT scan of the abdomen) is the most likely answer.} Option 'A' (Esophagogastroduodenoscopy) is not indicated as this patient doesn't show any alarming symptoms like weight loss, anemia, or dysphagia that are typical for upper GI malignancies. Option 'C' (Hydrogen breath test) is used to diagnose conditions like lactose intolerance and bacterial overgrowth, which are not likely in this case. \greenhighlight{Option 'D' (Cardiac stress test) is not necessary since there's no evidence of cardiac disease.} Option 'E' (Abdominal ultrasonography of the right upper quadrant) is usually used to evaluate hepatobiliary disease, which this patient doesn't seem to have. & The part highlighted in \greenhighlight{light green} refers to the portion where GPT-4 refutes the reference (ground-truth) answer, i.e. point ii.a of error category description. The part highlighted in \yellowhighlight{yellow} refers to the portion where GPT-4 explains its predicted option, i.e., point (ii.b) of the error category description. The lines highlighted in \textcolor{red}{red} contradict the point that there are strong indications or possibilities of cardiac disease.\\ \hline 
               \textbf{Error 6: Incorrect Understanding of Task} Question ID: \textit{23}, Correct Answer: (C) Delirium tremens & \greenhighlight{This patient's fever, tachycardia, hypertension, agitation, and visual hallucinations (garbage and flies in the room) in the setting of recent hospitalization and surgery suggest} delirium, a fluctuating disturbance in attention and cognition. \greenhighlight{This patient's delirium is most likely due to her infection and use of the antipseudomonal antibiotic piperacillin-tazobactam, which can cause neurologic side effects including confusion and hallucinations, especially in older patients and those with renal impairment} \yellowhighlight{(Option B)}. Acute cholangitis \yellowhighlight{(Option A)} usually presents with Charcot triad (fever, abdominal pain, jaundice) or Reynold pentad (Charcot triad plus hypotension and altered mental status). However, this patient has no abdominal pain and has been on appropriate antibiotics therapy for her infection. Alcohol hallucinosis \yellowhighlight{(Option C)} is a syndrome of sustained auditory hallucinations occurring in clear consciousness that develops within 48 hours of reduction or cessation of heavy alcohol intake. This patient does not have a reported history of alcohol abuse and her hallucinations are visual, not auditory. Delirium tremens \yellowhighlight{(Option D)} \greenhighlight{is a severe, life-threatening form of alcohol withdrawal that includes delirium, hallucinations, severe autonomic hyperactivity (eg, fever, tachycardia, hypertension), and seizures. However, it typically occurs 48-96 hours after the last drink. This patient does not have a reported history of alcohol abuse and her symptoms developed in the hospital, not after cessation of alcohol. Hepatic encephalopathy} \yellowhighlight{(Option D)} \greenhighlight{presents with confusion, asterixis, and hyperreflexia due to hyperammonemia in the setting of acute or chronic liver disease.} This patient has no known liver disease. Thyroid storm (Option E) is a life-threatening exacerbation of hyperthyroidism that usually occurs in patients with known disease. Manifestations include fever, tachycardia, hypertension, agitation, and delirium. However, other signs or symptoms of hyperthyroidism (eg, weight loss, heat intolerance, goiter, ophthalmopathy) are typically present, none of which are seen in this patient. &  GPT-4 refutes all the options. Although it initially selects option B, it later also refutes option B. Therefore, it did not follow the instruction ``Which of the following is the most likely diagnosis in this patient? Please explain your answer and why the other options are wrong.'' \\ \hline
               \textbf{Error 7: Hallucination of Information} Question ID: \textit{81}, Correct Answer: \textit{(C) 21-hydroxylase} & This patient's findings suggest 11ß-hydroxylase deficiency, a form of congenital adrenal hyperplasia (CAH). This condition results in excessive production of adrenal androgens, causing virilization (as evidenced by the presence of pubic hair) and tall stature in childhood (due to early epiphyseal plate closure). The deficiency also leads to overproduction of 11-deoxycorticosterone, a mineralocorticoid precursor, which leads to hypertension and hypokalemia (hence the high renin and potassium levels). Lastly, the high 17-hydroxyprogesterone is a common finding in all forms of CAH due to feedback upregulation of ACTH. Option A, 17a-hydroxylase deficiency, presents as female phenotype in both genetic males and females (due to lack of sex hormone production) and hypertension (due to increased mineralocorticoid production). Option C, 21-hydroxylase deficiency, is the most common form of CAH, which usually presents with salt wasting and virilization, \yellowhighlight{but not with hypertension}. Option D, Aromatase deficiency, results in masculinization of external genitalia in genetic females and tall stature in both males and females, but it does not cause hypertension. Option E, 5a-reductase deficiency, would result in ambiguous genitalia (i.e., male pseudohermaphroditism) in genetic males, but would not cause hypertension. & The ground-truth correct answer is C whereas GPT-4 selects option B. While explaining why Option C is incorrect, GPT-4 introduces a new symptom “hypertension” that is not explicitly mentioned in the USMLE question. The additional information is what we refer to as hallucination. \\ \hline
     \end{tabular}}
    \caption{Example and explanation of error classes of our proposed error taxonomy}
    \label{tab:pos-exam-reason-errors}
\end{table*}

\if{0}
\begin{table*}[!ht]
    \centering
    \scalebox{0.7}{
    \begin{tabular}{p{1.8cm}p{19cm}p{4cm}} \hline
          \textbf{Error Type}& \textbf{GPT-4 Response}&\textbf{Explanation of GPT-4 Response}\\ \hline
                  \textbf{Unsupported Medical Claim} [Question ID: X, Correct Answer: (D) Cardiac stress test] & This patient's abdominal discomfort with exertion and postprandial discomfort may suggest "abdominal angina" due to chronic mesenteric ischemia (CMI). CMI is most frequently observed in patients with generalized atherosclerotic cardiovascular disease which can be surmised from this patient's history of smoking, peripheral arterial disease, hypertension, and type 2 diabetes mellitus. The splanchnic circulation has a rich collateral flow. Consequently, at least two of the three major splanchnic vessels (celiac trunk, superior mesenteric artery, inferior mesenteric artery) must be significantly stenotic before symptoms of CMI develop. This patient's postprandial abdominal pain is due to an inability to augment splanchnic blood flow in response to increased metabolic demand after meals. As a result, patients may eat less and lose weight. The pain after a "big meal" and his inability to gain symptomatic relief despite a significant decrease in food intake due to fear of provoking the pain support this diagnosis. Atherosclerosis in other vascular beds can cause symptoms (eg, claudication), but this patient's discomfort with walking is more likely to be due to his deconditioning and obesity than to his stage 2 peripheral arterial disease, which is typically asymptomatic. Computed tomographic angiography (CTA) of the abdomen is the best initial diagnostic test for CMI. Therefore, Answer 'B' (CT scan of the abdomen) is the most likely answer. Option 'A' (Esophagogastroduodenoscopy) is not indicated as this patient doesn't show any alarming symptoms like weight loss, anemia, or dysphagia that are typical for upper GI malignancies. Option 'C' (Hydrogen breath test) is used to diagnose conditions like lactose intolerance and bacterial overgrowth, which are not likely in this case. Option 'D' (Cardiac stress test) is not necessary since there's no evidence of cardiac disease. Option 'E' (Abdominal ultrasonography of the right upper quadrant) is usually used to evaluate hepatobiliary disease, which this patient doesn't seem to have. & The part highlighted in light green refers to the portion where GPT-4 refutes the reference (ground-truth) answer, i.e. point ii.a of error category description. The part highlighted in yellow refers to the portion where GPT-4 explains its predicted option, i.e., point (ii.b) of the error category description. The lines highlighted in red contradict the point that there are strong indications or possibilities of cardiac disease.\\ \hline 
        \textbf{Hallucination of Information} [Question ID: X, Correct Answer: \textit{(C) 21-hydroxylase}] & This patient's findings suggest 11ß-hydroxylase deficiency, a form of congenital adrenal hyperplasia (CAH). This condition results in excessive production of adrenal androgens, causing virilization (as evidenced by the presence of pubic hair) and tall stature in childhood (due to early epiphyseal plate closure). The deficiency also leads to overproduction of 11-deoxycorticosterone, a mineralocorticoid precursor, which leads to hypertension and hypokalemia (hence the high renin and potassium levels). Lastly, the high 17-hydroxyprogesterone is a common finding in all forms of CAH due to feedback upregulation of ACTH. Option A, 17a-hydroxylase deficiency, presents as female phenotype in both genetic males and females (due to lack of sex hormone production) and hypertension (due to increased mineralocorticoid production). Option C, 21-hydroxylase deficiency, is the most common form of CAH, which usually presents with salt wasting and virilization, but not with hypertension. Option D, Aromatase deficiency, results in masculinization of external genitalia in genetic females and tall stature in both males and females, but it does not cause hypertension. Option E, 5a-reductase deficiency, would result in ambiguous genitalia (i.e., male pseudohermaphroditism) in genetic males, but would not cause hypertension. & The ground-truth correct answer is C whereas GPT-4 selects option B. While explaining why Option C is incorrect, GPT-4 introduces a new symptom “hypertension” that is not explicitly mentioned in the USMLE question. The additional information is what we refer to as hallucination. \\ \hline
        \textbf{Incorrect Understanding of Task} [Question ID: X, Correct Answer: (C) Delirium tremens] & This patient's fever, tachycardia, hypertension, agitation, and visual hallucinations (garbage and flies in the room) in the setting of recent hospitalization and surgery suggest delirium, a fluctuating disturbance in attention and cognition. This patient's delirium is most likely due to her infection and use of the antipseudomonal antibiotic piperacillin-tazobactam, which can cause neurologic side effects including confusion and hallucinations, especially in older patients and those with renal impairment (Option B). Acute cholangitis (Option A) usually presents with Charcot triad (fever, abdominal pain, jaundice) or Reynold pentad (Charcot triad plus hypotension and altered mental status). However, this patient has no abdominal pain and has been on appropriate antibiotics therapy for her infection. Alcohol hallucinosis (Option C) is a syndrome of sustained auditory hallucinations occurring in clear consciousness that develops within 48 hours of reduction or cessation of heavy alcohol intake. This patient does not have a reported history of alcohol abuse and her hallucinations are visual, not auditory. Delirium tremens (Option D) is a severe, life-threatening form of alcohol withdrawal that includes delirium, hallucinations, severe autonomic hyperactivity (eg, fever, tachycardia, hypertension), and seizures. However, it typically occurs 48-96 hours after the last drink. This patient does not have a reported history of alcohol abuse and her symptoms developed in the hospital, not after cessation of alcohol. Hepatic encephalopathy (Option D) presents with confusion, asterixis, and hyperreflexia due to hyperammonemia in the setting of acute or chronic liver disease. This patient has no known liver disease. Thyroid storm (Option E) is a life-threatening exacerbation of hyperthyroidism that usually occurs in patients with known disease. Manifestations include fever, tachycardia, hypertension, agitation, and delirium. However, other signs or symptoms of hyperthyroidism (eg, weight loss, heat intolerance, goiter, ophthalmopathy) are typically present, none of which are seen in this patient. &  GPT-4 refutes all the options. Although it initially selects option B, it later also refutes option B. Therefore, it did not follow the instruction ``Which of the following is the most likely diagnosis in this patient? Please explain your answer and why the other options are wrong.'' \\ \hline

    \end{tabular}}
    \caption{Example and explanation of one knowledge-based error category of ``Unsupported medical claim'', followed by reading comprehension error categories, developed as part of the annotation guidelines document \notesr{Please highlight same as shown in the annotation guidelines document. Use the following Latex text highlighting commands \greenhighlight{span1} \yellowhighlight{span2} }}
    \label{tab:pos-exam-compre-errors}
\end{table*}
\fi

\subsubsection{Non-error types}
We further define two categories that can be chosen if the annotator feels that none of the previously described error types are applicable:

\noindent \textbf{Non-error 1: Reasonable response by GPT-4.} 
%You can select this option if you find the explanation provided by GPT-4 reasonable and the explanation does not have incorrect facts or flawed reasoning. Since GPT-4 selects the incorrect option, the answer by GPT-4 is also a potential or more medically reasonable answer.
This option can be selected if the explanation provided by GPT-4 is found reasonable, and the explanation does not contain incorrect facts or flawed reasoning. Given that the incorrect option is chosen by GPT-4, the response is still considered a medically reasonable answer. %\notesr{We can add the example from the annotation guidelines, with the highlighted portions plus an explanation for choosing them}

\noindent \textbf{Non-error 2: Cannot pick any category.} 
%You can select this option when you find certain flaws in the facts or reasoning of GPT-4 and are not fully convinced with the GPT-4 response. However, you are also not able to assign the GPT-4 response to any of the error categories.
This option can be chosen when certain flaws in the facts or reasoning of GPT-4 are identified, and when one is not fully convinced with the GPT-4 response. However, the ability to assign the GPT-4 response to any of the error categories is also not possible.

\subsection{Annotator Details}\label{sec:annotator-details}
According to previous research, a majority vote among annotators is commonly employed to establish ground truth labels in NLP tasks. However, discrepancies in labels, particularly in tasks like hate speech detection, may indicate systematic variations in opinion due to differences in annotators' beliefs \cite{fleisig2023majority,wan2023everyone,pei2023annotator}. These systematic differences become even more crucial in the medical domain, given the divergent training and study approaches influenced by country-specific guidelines and medical practices \cite{harrigian2023characterization}. To delve into this distribution, we conducted an exploratory study of annotators, considering their demographic profiles.

\subsubsection{Demographics} 
We gathered annotations from 44 participants (28 female) of 17 nationalities, living in 11 countries, with most (n=26) living in the USA or UK. The participants were between 24 and 71 years of age, with 75\% of participants aged 35 or younger. All were required to speak English fluently and have completed a graduate degree (n=38) or higher in Biomedical Science, Medicine, or Health and Medicine. Of these, 10 participants stated they were students pursuing another degree, 33 were employed and one participant was self-employed. The participants are highly experienced with a mean of 553 approved submissions on Prolific. We further analyze the medical expertise of the annotators and observe that a very high percentage of them ($75\%$) are currently practicing or previously practiced in clinical settings. Additionally, a good amount of annotators ($54.\overline{54}\%$) have taken or passed the USMLE exam, while $61.\overline{36}\%$ were already familiar with the format. %However, we also observed that only half of the annotators had worked with AI-based models in a medical context.

\subsubsection{Annotation Experience} Participants were very happy with both the annotation platform as well as the study overall. More than 75\% felt that the annotation guidelines were satisfactory or better, while more than 90\% and 98\% felt satisfied or better about their experience participating in the study and the annotation platform, respectively. We also collected textual feedback and suggestions for improvement during our study (15 out of 44 responded) by asking ``\textit{Please feel free to leave any comments about our study. If you give a poor score on any of the above points, please write in detail about the issue and suggestions for improvements}''. While $40\%$ of the feedback was positive, three suggestions were mentioned most: (i) Three participants mentioned that our study was very time-consuming and wished for a timer to be displayed or a bonus component for sincere effort, (ii) four annotators provided suggestions on how to improve the annotation guidelines, e.g. summarizing and upping the audio of the accompanying video and (iii) two participants called the ground truth answer into question, arguing another answer should be correct or the GPT-4 answer was not outright wrong, but less correct than the ground truth.

\subsection{Finalizing the Annotations}\label{sec:final-annot}

%Let us take the example from Table~\ref{tab:multi-label-example} where 
%Suppose annotator A annotates a data point as \textit{{`Sticking with the wrong diagnosis', `Hallucination of information'}}. We represent each data point as a set of nine elements, corresponding to the nine class labels, for computing the Jaccard similarity score: \textit{[`Sticking with the wrong diagnosis positive', `Hallucination of information positive', `Incorrect or vague conclusion negative', `Unsupported medical claim negative', 'Incorrect understanding of the task negative', `Cannot pick any error category negative', `Reasonable response by GPT-4 negative', 'Ignore missing information negative', 'Non-medical factual error negative']}. 

\begin{figure}[t]
    \centering
    \includegraphics[width=0.4\textwidth]{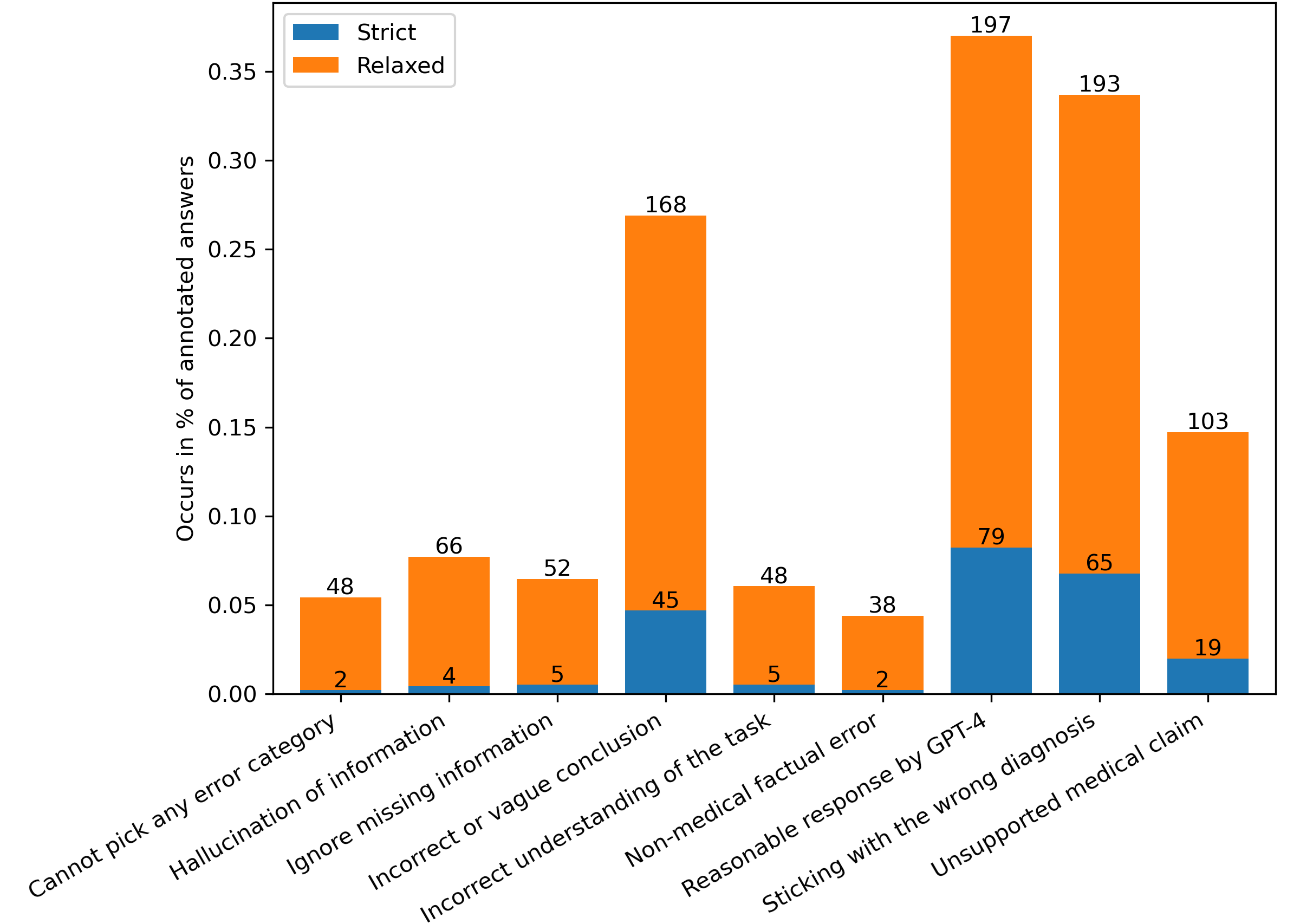}
    \caption{Label distribution of multi-label \gue{} dataset}
    \label{fig:data-stats-multi-label}
\end{figure}

We only consider labels that are given by at least two annotators because it makes the labels more likely to represent the GPT-4 Response; we follow the same procedure as used by ~\citet{sohamcaves2022} for finalizing the multi-label annotations. Suppose a GPT-4 response is labeled by three annotators (A1 - A3) as follows --- A1: {C1, C2, C3}; A2: {C2, C3, C4}; A3: {C3}, where C1 - C4 represents some classes, then the final ground-truth labels will be L: {C2, C3}. 
Because these labels are chosen by at least two annotators, an approach commonly utilized in other studies within the medical domain \cite{dash2023evaluation}. The number of data points that satisfy these criteria is 191 out of 300. We refer to it as the \textbf{strict} criterion The annotation agreement score for the data points selected using the `strict' criterion (191 data points), computed in terms of pair-wise Jaccard similarity score is $0.7 \pm 0.15$ (mean) and $0.7$ (median), which is an increase of $6.06\%$ and $9.38\%$ respectively over the entire annotated dataset of 300 data points. We note that $28.78\%$ of answers were labeled as reasonable responses, while the error classes 1, 2, and 5 %\textit{sticking with wrong diagnosis}, \textit{incorrect or vague conclusion} and \textit{unsupported medical claim}
occur the most ($61.83\%$ in total). We also provide statistics for the \textbf{relaxed} criterion, where we consider all labels given by any of the annotators. Figure~\ref{fig:data-stats-multi-label} shows the distribution of the labels for both the strict and relaxed case, displaying both the total number for each error class as well as the percentage of annotated questions it was labeled.

We now compute the inter-annotator agreement statistics for the complete dataset of 300 points, where each data point is a multi-label annotation.  Given the multi-label nature of the dataset with several sets of annotators, we could not directly use Cohen or Fleiss kappa for measuring the agreement. Therefore, we use the Jaccard coefficient, an overlap-based metric. We first compute the mean pair-wise Jaccard similarity score (overlap of annotation labels - both positive and negative) for each data point. We observe that the mean and median agreement statistics are quite high, with values of $66 \pm 14\%$ and $64\%$ respectively, which indicates a reasonably high degree of agreement.

\begin{table}[t]
    \centering
    \scalebox{0.7}{
    \begin{tabular}{|p{10cm}|} \hline
        \textbf{Question id: } 9499  \\
        \textbf{Correct Answer: } (D) The gene locus which is mutated in this condition is on chromosome 9 \\ \hline \hline
         \textbf{GPT-4 Response: } The correct statement related to the diagnosis of this patient is: 'C: Gait ataxia in this condition is a pure sensory ataxia'. The patient's symptoms and the result of the genetic testing indicate that he has Friedreich's ataxia, a rare genetic disease that causes difficulty walking, a loss of sensation in the arms and legs, and impaired speech. This condition is caused by \textit{[...]} %This condition is caused by an expansion of GAA trinucleotide repeats in the frataxin gene on chromosome 9. The disease is inherited in an autosomal recessive pattern, not dominant,
         which eliminates option 'E'. Patients with Friedreich's ataxia usually present with gait ataxia, \textit{[...]} %which is primarily of sensory origin due to the degeneration of the dorsal columns and spinocerebellar tracts. Hence, the ataxia in this condition can be correctly referred to as a 'pure sensory ataxia'.
         Option 'A' is wrong because horizontal, not vertical, nystagmus is typically seen in patients with this condition. Option 'B' is incorrect because although heart disease is a common cause of death in those with Friedreich's ataxia, it is hypertrophic cardiomyopathy, \greenhighlight{not restrictive cardiomyopathy, that is commonly seen. Finally, option 'D' is wrong because the mutated gene in Friedreich's ataxia is} \yellowhighlight{located on chromosome 9q13-q21.1, not just chromosome 9}. So, its location is more specific than merely chromosome 9.\\ \hline \hline
         \textbf{Error Annotation: } \greenhighlight{Sticking with the wrong diagnosis}, \yellowhighlight{Hallucination of information}\\ \hline
    \end{tabular}}
    \caption{Example of a multi-label span annotation}
    \label{tab:multi-label-example}
\end{table}

%To check the \textit{Inter Annotator Agreement} (IAA), we calculated \textit{Krippendorff's alpha agreement} score (using MASI distance \notesr{Add citation from CAVES paper Section 3.4 third paragraph}). The agreement score obtained is X \notesr{Needs to be computed} on the final dataset (considering only the labels assigned by at least two annotators - the ``strict'' condition.

%\subsubsection{Finalizing the explanation spans}
%We follow the strategy used by Lee et al.~\cite{Lee2019} for finalizing the explanation spans.

\subsection{Additional Resources}\label{sec:add-resources}

\subsubsection{Medical Concepts and Predicates Extraction with SemRep~\cite{Kilicoglu2020, rindflesch2003interaction}}\label{sec:semrep-analysis}
\begin{figure}[!ht]
    \centering
    \includegraphics[width=0.3\textwidth]{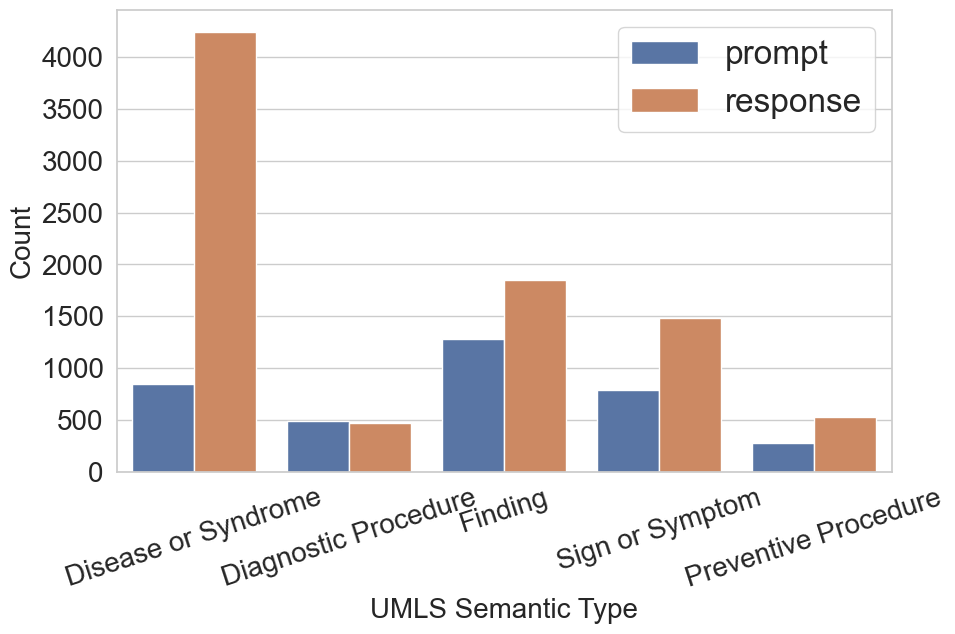}
    \hfill
    \includegraphics[width=0.3\textwidth]{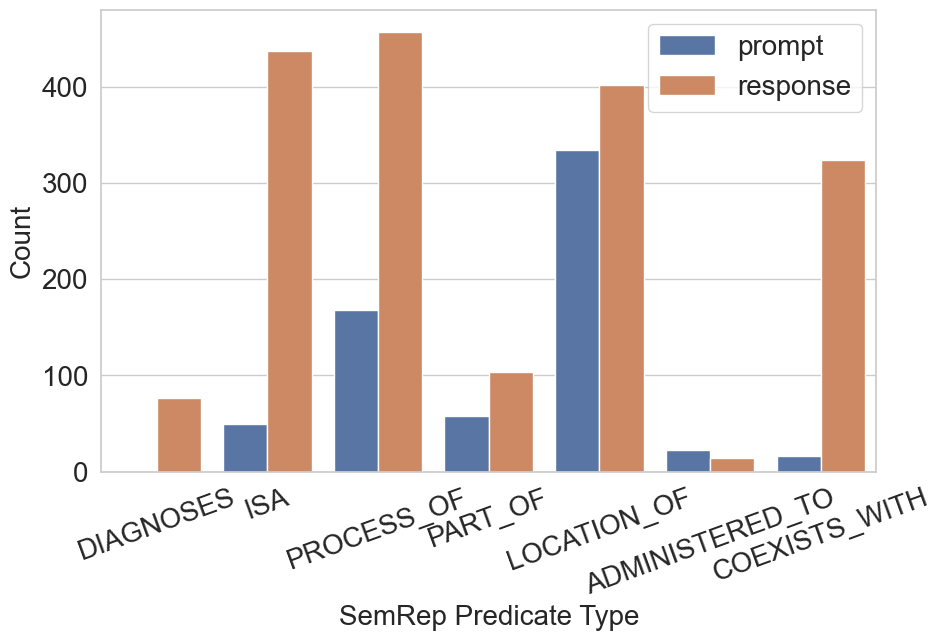}
    \caption{Distribution of UMLS Semantic Types of medical concepts (top) and Predicate Types of UMLS Predications (bottom) extracted using the SemRep tool}
    \label{fig:dist-semrep-ent-pred}
\end{figure}

Here, we aim to identify medical phrases from the medical text (both USMLE question and GPT-4 Response). %The MedQA paper~\cite{jin2021disease} provides only medical phrases extracted using MetaMap but does not provide the associated Unified Medical Language System\footnote{\url{https://www.nlm.nih.gov/research/umls/index.html}} (UMLS) concept id, semantic type, or semantic predications.
SemRep is used to extract semantic predications, consisting of a subject argument, an object argument, and the relation that binds them, based on the Unified Medical Language System\footnote{\url{https://www.nlm.nih.gov/research/umls/index.html}} (UMLS). The subject and object arguments of each prediction are concepts from the UMLS Methathesaurus, while their relation stems from the UMLS Semantic Network~\cite{Kilicoglu2020, rindflesch2003interaction}. Table~\ref{tab:semrep-example} provides an example for our G4UE dataset, where we show the semantic predications extracted from a single sentence. We use the \textit{Batch SemRep} online tool\footnote{\url{https://ii.nlm.nih.gov/Batch/UTS_Required/SemRep.html}} to extract UMLS concept ids, its associated semantic type, and semantic predications from both the USMLE question and response from GPT-4.
 %For a detailed description of SemRep, see ~\cite{Kilicoglu2020, rindflesch2003interaction}.
%\tb{SemRep is a UMLS-based program that extracts three-part propositions, called semantic predications, from sentences in biomedical text. Predications consist of a subject argument, an object argument, and the relation that binds them. The subject and object arguments of each predication are concepts from the UMLS Metathesaurus and their binding relationship (in uppercase) is a relation from the UMLS Semantic Network. For a detailed description of SemRep, see~\cite{Kilicoglu2020,rindflesch2003interaction}}.\notesr{please paraphrase}
 We use the following parameters while using the Batch SemRep tool --- \textit{Anaphora Resolution} (-A), Knowledge Source (-Z) as \textit{2018}, Lexicon Year (-L) as 2018, Data Model as \textit{Strict Model}, and output should be in XML format (-X). Figure~\ref{fig:dist-semrep-ent-pred} shows the distribution of UMLS Semantic Types (based on UMLS concept IDs) and semantic predications extracted using the SemRep tool.

%as described in detail in Section~\ref{sec:semrep-analysis}. \notesr{what is the inference?}
\begin{table}[!ht]
    
    \centering
    \scalebox{0.7}{
    \begin{tabular}{|p{0.4cm}p{1cm}p{2.5cm}p{2.4cm}p{2.4cm}|} \hline
        \multicolumn{5}{|c|}{\textbf{\textsc{Sentence}}}\\
         \multicolumn{5}{|p{10cm}|}{Duplex ultrasound shows vascular wall thickening and subcutaneous edema.}\\ \hline \hline
         \multicolumn{5}{|c|}{\textbf{\textsc{Medical Entities}}}\\
         \textbf{Id}&  \textbf{UMLS Concept Id}&  \textbf{UMLS Concept Name}&  \textbf{UMLS Semantic Type}& \textbf{Text Segment}\\ \hline
         Ent1&  C0242845&  Ultrasonography,
Doppler, Duplex&  Diagnostic Procedure& Duplex ultrasound\\
         Ent2&  C118003&  Wall of blood vessel&  Body Part, Organ, or Organ Component& vascular wall\\
         Ent3&  C0205400&  Thickened&  Finding& thickening\\
 Ent4& C0241277& Swelling of subcutaneous tissue& Finding&subcutaneous edema\\ \hline \hline
 \multicolumn{5}{|c|}{\textbf{\textsc{Medical Predications using SemRep}}}\\ 
 \textbf{Id}& \multicolumn{2}{c}{\textbf{Subject Entity Id}}& \textbf{Predicate Type}& \textbf{Object Entity Id}\\ \hline
 Pred1& \multicolumn{2}{c}{Ent2}& Location of& Ent3\\
 Pred2& \multicolumn{2}{c}{Ent1}& Diagnoses& Ent3\\
 Pred3& \multicolumn{2}{c}{Ent1}& Diagnoses& Ent4\\ \hline
    \end{tabular}}
    \caption{Medical concepts and predications extracted using SemRep tool for a sentence in USMLE question}
    \label{tab:semrep-example}
\end{table}

\subsubsection{GPT-4 Responses taken on January 28, 2024}
We provide GPT-4 responses to the 919 data points where it selected the incorrect option, at two time points - (i) August 30, 2023, and (ii) January 28, 2024. Since we not only provide the predicted option by GPT-4 but also provide the long-form explanation both for selecting the correct option and for refuting the other options, this proves to be a rich resource for performing in-depth analysis of drift behavior~\cite{chen2023chatgpt}. 

\if{0}
\begin{table}[!ht]
\scriptsize
    \centering
    \begin{tabular}{p{8cm}} \hline 
  \textbf{Sentence:} Duplex ultrasound shows vascular wall thickening and subcutaneous edema. \\ \hline
  \textbf{Entities:} 
  Entity id=``E1", UMLS concept id (cui)=``C0242845" cui name=``Ultrasonography, Doppler, Duplex" semtypes=``Diagnostic Procedure", text="Duplex ultrasound" />
  
  <Entity id=``Dtext\_000N\_16280.E24" cui=``C1180033" name=``Wall of blood vessel" semtypes=``Body Part, Organ, or Organ Component'' text=``vascular wall" />
  
  <Entity id=``Dtext\_000N\_16280.E25" cui=``C0205400" name=``Thickened" semtypes=``Finding" text="thickening" />
  
  <Entity id=``Dtext\_000N\_16280.E26" cui=``C0241277" name=``Swelling of subcutaneous tissue" semtypes=``Finding" text="subcutaneous edema" /> \\ \hline
  
  \textbf{Predications:}  <Predication id="Dtex\_000N\_16280.P2" >
   <Subject entityID="Dtext\_000N\_16280.E24" relSemType=``Body Part, Organ, or Organ Component" /> <Predicate type="LOCATION\_OF" indicatorType="MOD\_HEAD"/>   <Object entityID="Dtext\_000N\_16280.E25" relSemType="Finding" />  </Predication>
  
  <Predication id=``Dtext\_000N\_16280.P3">
   <Subject entityID="Dtext\_000N\_16280.E23" relSemType="Diagnostic Procedure" />
   <Predicate type="DIAGNOSES" indicatorType="VERB" />
   <Object entityID="Dtext\_000N\_16280.E25" relSemType=``Finding" />  </Predication>
  
  <Predication id=``Dtext\_000N\_16280.P4" >
   <Subject entityID="Dtext\_000N\_16280.E23" relSemType="Diagnostic Procedure" />
   <Predicate type="DIAGNOSES" indicatorType="VERB" />
   <Object entityID=``Dtext\_000N\_16280.E26" relSemType=``Finding" />
  </Predication> \\ \hline    
    \end{tabular}
    \caption{Medical concepts and predications extracted using SemRep tool for a sentence in USMLE question \notesr{simplify and remove junk text}}
    \label{tab:semrep-example}
\end{table}
\fi

\if{0}
\begin{table}[!ht]
\scriptsize
    \centering
    \begin{tabular}{p{8cm}} \hline 
  \textbf{Sentence:} Duplex ultrasound shows vascular wall thickening and subcutaneous edema. \\ \hline
  \textbf{Entities:} 
  <Entity id=``Dtext\_000N\_16280.E23" cui=``C0242845" name=``Ultrasonography, Doppler, Duplex" semtypes=``Diagnostic Procedure", text="Duplex ultrasound" />
  
  <Entity id=``Dtext\_000N\_16280.E24" cui=``C1180033" name=``Wall of blood vessel" semtypes=``Body Part, Organ, or Organ Component'' text=``vascular wall" />
  
  <Entity id=``Dtext\_000N\_16280.E25" cui=``C0205400" name=``Thickened" semtypes=``Finding" text="thickening" />
  
  <Entity id=``Dtext\_000N\_16280.E26" cui=``C0241277" name=``Swelling of subcutaneous tissue" semtypes=``Finding" text="subcutaneous edema" /> \\ \hline
  
  \textbf{Predications:}  <Predication id="Dtex\_000N\_16280.P2" >
   <Subject entityID="Dtext\_000N\_16280.E24" relSemType=``Body Part, Organ, or Organ Component" /> <Predicate type="LOCATION\_OF" indicatorType="MOD\_HEAD"/>   <Object entityID="Dtext\_000N\_16280.E25" relSemType="Finding" />  </Predication>
  
  <Predication id=``Dtext\_000N\_16280.P3">
   <Subject entityID="Dtext\_000N\_16280.E23" relSemType="Diagnostic Procedure" />
   <Predicate type="DIAGNOSES" indicatorType="VERB" />
   <Object entityID="Dtext\_000N\_16280.E25" relSemType=``Finding" />  </Predication>
  
  <Predication id=``Dtext\_000N\_16280.P4" >
   <Subject entityID="Dtext\_000N\_16280.E23" relSemType="Diagnostic Procedure" />
   <Predicate type="DIAGNOSES" indicatorType="VERB" />
   <Object entityID=``Dtext\_000N\_16280.E26" relSemType=``Finding" />
  </Predication> \\ \hline    
    \end{tabular}
    \caption{Medical concepts and predications extracted using SemRep tool for a sentence in USMLE question \notesr{simplify and remove junk text}}
    \label{tab:semrep-example}
\end{table}
 \fi

\if{0}
\subsection{Additional resources for \gue{} dataset}
We also make the remaining wrong responses publicly available ($619$ out of $919$ wrong responses are still unannotated) and the correct responses to the USMLE questions, which is around $4153$ data points.
\fi

\section{Characterization Study}
  Here, we perform an in-depth characterization study of the error classes and utilize the resources that we contribute in this work to draw interesting insights. Similar to the work of Lievin et al.~\cite{liévin2023large} we find that most incorrect answers in the medical domain are reasoning-based and note a rather low amount of factual errors as seen in the software engineering domain before~\cite{kabir2023answers}. 

\subsection{Error Co-occurrence Details}
\begin{figure}[t]
    \centering
    \includegraphics[width=0.47\textwidth]{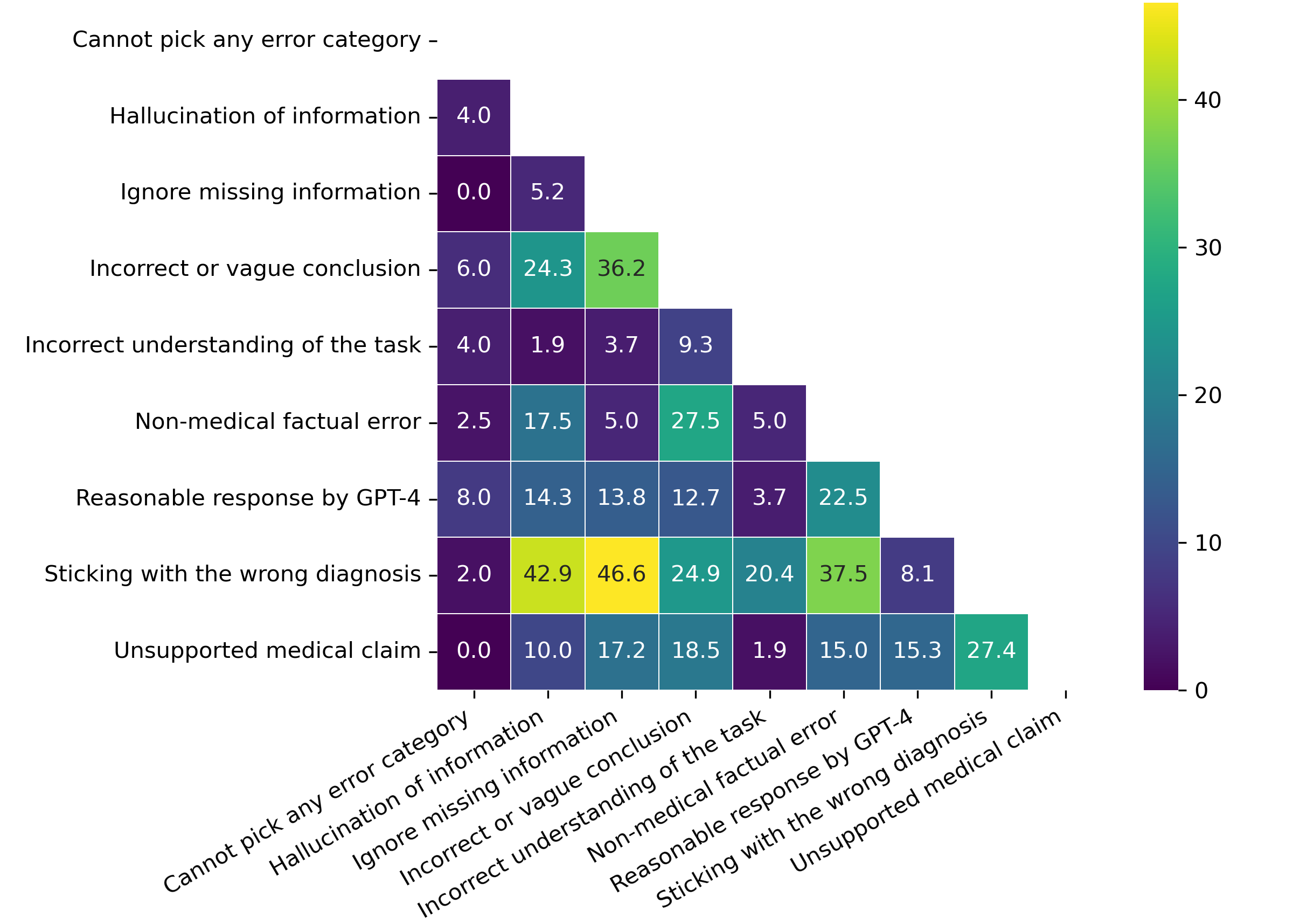}
    \caption{Error class co-occurrence statistics}
    \label{fig:error-type-corr}
\end{figure}

 Figure~\ref{fig:error-type-corr} shows the joint distribution of the error classes and helps to get a better understanding of the interplay between the error classes. To do this, we divide the number of answers that have been labeled with any two error classes $c_i$ and $c_j$, $n_{ij}$, by how often both classes could have occurred together (i.e. $min\{|c_i|,|c_j|\}$) and multiply by 100 for scaling. We note that every error type (i.e. excluding the non-error classes) except \textit{error 4} co-occurs most often with \textit{error 1}. 

\subsection{Analysis of Reasoning Errors}
\label{sec:reasoning-based-error}
We recognize the three reasoning-based error types constitute the majority of non-fact-check-able errors, with \textit{Sticking with the wrong diagnosis} 
($26.9\%$) and \textit{Incorrect or vague conclusion}
($22.21\%$) being the most common errors overall. This is similar to the trend observed by a recent study~\cite{liévin2023large} for GPT-3.5; however, we provide a more fine-grained error taxonomy for GPT-4. 

We now analyze the most common error (\textbf{sticking with the wrong diagnosis}) and its co-occurrence patterns, and observe that it largely occurs in combination with another error class ($64.77\%$ of cases). It occurs before the other error(s) much more often than afterward ($60\%$ of cases compared to $26.4\%$). When labeled with error classes that co-occurs quite frequently ($n_{ij}<.33$, see Figure~\ref{fig:error-type-corr}), it appears first much more often ($\ge 73.\overline{33}\%$ of cases). A deeper look into a subset of these cases reveals that once GPT-4 has made a choice (be it choosing something as a correct option or making a diagnosis), it will henceforth treat this as the truth and will try to justify this `truth' by any means. We note two typical behaviors: (i) for the justification to sound reasonable, GPT-4 will hallucinate, ignore that crucial information is missing, or make factual errors or (ii) it will continue with self-contradictory reasoning, i.e. provide sound explanations for the other options, yet refrain from correcting the earlier answer. As such, we posit that some of the other (highly co-occurring) errors only materialize due to a prior reasoning misstep and highlight the importance of studying this behavior further.

\begin{table}[t]
%\scriptsize
    \centering
    \scalebox{0.85}{
    \begin{tabular}{lccc} \hline
        \textbf{Co-occurring error type} & \textbf{First} & \textbf{Middle} & \textbf{Last} \\
        \midrule
        Hallucination of information$^\dagger$ & $76.\overline{66}$\% & $0.00\%$ & $23.\overline{33}\%$ \\
        Ignore missing information$^\dagger$ & 7$4.07\%$ & $0.00\%$ & $25.93\%$ \\
        Incorrect or vague conclusion & $60.00\%$ & $3.\overline{63}\%$ & $36.\overline{36}\%$ \\
        Incorrect understanding of the task & $45.\overline{45}\%$ & $9.\overline{09}\%$ & $45.\overline{45}\%$ \\
        Non-medical factual error$^\dagger$ & $73.\overline{33}\%$ & $0.00\%$ & $26.\overline{67}\%$ \\
        Unsupported medical claim & $67.65\%$ & $2.94\%$ & $29.41\%$ \\
        \midrule
        Average position overall & $60.00\%$ & $13.60\%$ & $26.40\%$ \\
        \bottomrule
    \end{tabular}}
    \caption{Positional statistics of \textit{sticking with the wrong diagnosis} in answers with multiple error labels. \textit{Middle} means that multiple errors are annotated before and after it. $^\dagger$ indicates frequently co-occurring error classes}
    \label{tab:sticking-error-positional}
\end{table}

\subsection{Understanding Drift Behavior of GPT-4}
We perform an initial comparison of GPT-4 long-form responses between the two time points of August 2023 and January 2024, to study the drift behavior of GPT-4. We observe that for 214 out of 919 responses, GPT-4 now selects the correct option. Next, we perform an error-type-level analysis using only the developed \gue{} dataset (71 out of the 300 data points) keeping the \textit{strict} criteria; this criterion reduced it from 71 to 47 data points. The highest error types that showed positive drift behavior are \textit{non-error 1, error 1} and \textit{error 2} for $27$, $13$, and $8$ data points respectively. This further highlights the harmful and convincing nature of GPT-4 responses, where even when it selects the wrong option, the medical experts find it reasonable, but after model updates, GPT-4 corrects itself to provide the right answer. We gave this set of points (27 in total) to our medical experts and asked them to provide their perspectives.
\section{Additional Tasks}
In addition to the subjects we discussed thus far, we outline some potential uses for the dataset in this section, showcasing the value our dataset contributes to the community. 

\noindent \textbf{Multi-label classification: document and span level.} In the multi-label classification task, each data point (here: GPT-4 response to USMLE question) has to be assigned to one or more error classes. This can help reduce the amount of undetected wrong answers by LLMs in the medical domain, moving one step closer to supportive operations in clinics. Given a model with good performance, our dataset could be used to compare the characteristics of different LLMs in the medical domain beyond measuring accuracy. The dataset can also be used to train a model for the automated evaluation of LLM rationales or explanations. This task could also be executed in a more detailed manner by conceptualizing it as a sentence- or multi-span question-answering task \cite{zhu2020question}.

\noindent \textbf{Impact of Medical Concepts.} The QA-pairs of the MedQA (and our) dataset are split into three \textit{steps}, testing either memorization or reasoning skills. We extracted relevant UMLS concepts and predicates using SemRep. Using this information, QA-pairs can be linked to more fine-grained topics, allowing further research that could yield useful insights into which medical areas GPT-4 shows better performance, such as enriching medical knowledge graph \cite{schutte2021discovering,li2021construction}. 

\noindent \textbf{Mitigating Reasoning-based Errors in LLMs.}
The performance of LLMs in clinical reasoning tasks has been a subject of investigation. GPT-3.5 struggled with advanced clinical reasoning tasks but GPT-4 showed improvement, yet not reaching human-like diagnostic accuracy. Despite this, GPT-4's ability to mimic clinical reasoning processes offers potential for interpretability, allowing clinicians to assess answers based on factual and logical accuracy, potentially mitigating "black box" limitations of language models \cite{savage2023diagnostic}. An interesting research direction will be to explore various strategies to reduce the amount of reasoning-based errors LLMs make. With methods like retrieval augmented generation and as evidenced by the low amount of knowledge-based errors in our study, the number of factual errors that current LLMs make is sinking. 

It is well-known that chain-of-thought prompting can alleviate some reasoning-related issues in LLMs~\cite{wei2022chain,Singhal2023}. Yet, further research regarding LLMs and their capability to reason is needed, as reasoning-based errors can be tough to detect and well-written texts can even fool professionals in complex domains such as medicine. Different reasoning frameworks like graph-of-thoughts~\cite{besta2023graph}, show that more reasoning can be inserted into LLMs at least extrinsically. We may consider the set of 4193 GPT-4 responses where it has selected the correct option as a training set containing correct reasoning examples, to teach the LLMs. In-context learning may be a possible research direction as it has shown success in improving the reasoning capabilities of LLMs~\cite{zhou2022teaching} where a subset of in-context examples are selected from this pool of positive GPT-4 responses and added as part of the prompt.

\noindent \textbf{Performance comparison with open-source LLMs.} We can generate responses from open-source LLMs on the full dataset of 5072 data points: (i) non-medical LLMs: Llama-2~\cite{touvron2023llama} 7B and 13B, and (ii) medical LLMs: MedAlpaca 7B~\cite{han2023medalpaca}. This resource may prove useful to perform in-depth comparisons of generated text quality instead of simply checking accuracy. These models have the added advantage that they do not suffer from \textit{performance drift}.

\section{Conclusion}\label{sec:conclu}
We introduce a new domain-specific error taxonomy and the GPT-4 USMLE Error dataset, a result of collaboration with medical students, provides a comprehensive resource for understanding the strengths and weaknesses of GPT-4 in medical QA tasks. Our large-scale annotation study, involving 44 medical experts, has shed light on the challenges of discerning explanations that may lead to incorrect options, even among trained professionals. The detailed explanations from GPT-4, along with the medical concepts and semantic predications provided for each question, offer valuable insights into the reasoning process of LLMs. We believe that these resources, available at \url{https://github.com/roysoumya/usmle-gpt4-error-taxonomy}, will significantly contribute to the evaluation and improvement of LLMs in answering complex medical questions. 

\noindent \textbf{Limitations.}  We require a better domain-specific method to handle the span annotations due to the complexity of medical text and inference. This necessitates considering relaxed boundaries for span selection, acknowledging the complexity of medical knowledge analysis. We use the Prolific crowdsourcing platform to recruit medical experts from a wide range of countries and may differ widely in terms of their reflection of medical knowledge and inference based on the annotator's country-specific medical training. Although we enforce multiple filters and quality check mechanisms, the task complexity and long annotation time may introduce fatigue in annotators, which in turn, may affect the quality of annotations. \citet{jin2024hidden} observed that the GPT-4 Vision model frequently generated incorrect explanations or rationales ($27.3\%$ of cases) even when it predicted the correct option. However, we only annotated the GPT-4 responses that predicted the wrong option in our study.

\if{0}
\section{Introduction}
ACM's consolidated article template, introduced in 2017, provides a
consistent \LaTeX\ style for use across ACM publications, and
incorporates accessibility and metadata-extraction functionality
necessary for future Digital Library endeavors. Numerous ACM and
SIG-specific \LaTeX\ templates have been examined, and their unique
features incorporated into this single new template.

If you are new to publishing with ACM, this document is a valuable
guide to the process of preparing your work for publication. If you
have published with ACM before, this document provides insight and
instruction into more recent changes to the article template.

The ``\verb|acmart|'' document class can be used to prepare articles
for any ACM publication --- conference or journal, and for any stage
of publication, from review to final ``camera-ready'' copy, to the
author's own version, with {\itshape very} few changes to the source.

\section{Template Overview}
As noted in the introduction, the ``\verb|acmart|'' document class can
be used to prepare many different kinds of documentation --- a
dual-anonymous initial submission of a full-length technical paper, a
two-page SIGGRAPH Emerging Technologies abstract, a ``camera-ready''
journal article, a SIGCHI Extended Abstract, and more --- all by
selecting the appropriate {\itshape template style} and {\itshape
  template parameters}.

This document will explain the major features of the document
class. For further information, the {\itshape \LaTeX\ User's Guide} is
available from
\url{https://www.acm.org/publications/proceedings-template}.

\subsection{Template Styles}

The primary parameter given to the ``\verb|acmart|'' document class is
the {\itshape template style} which corresponds to the kind of publication
or SIG publishing the work. This parameter is enclosed in square
brackets and is a part of the {\verb|documentclass|} command:
\begin{verbatim}
  \documentclass[STYLE]{acmart}
\end{verbatim}

Journals use one of three template styles. All but three ACM journals
use the {\verb|acmsmall|} template style:
\begin{itemize}
\item {\verb|acmsmall|}: The default journal template style.
\item {\verb|acmlarge|}: Used by JOCCH and TAP.
\item {\verb|acmtog|}: Used by TOG.
\end{itemize}

The majority of conference proceedings documentation will use the {\verb|acmconf|} template style.
\begin{itemize}
\item {\verb|acmconf|}: The default proceedings template style.
\item{\verb|sigchi|}: Used for SIGCHI conference articles.
\item{\verb|sigchi-a|}: Used for SIGCHI ``Extended Abstract'' articles.
\item{\verb|sigplan|}: Used for SIGPLAN conference articles.
\end{itemize}

\subsection{Template Parameters}

In addition to specifying the {\itshape template style} to be used in
formatting your work, there are a number of {\itshape template parameters}
which modify some part of the applied template style. A complete list
of these parameters can be found in the {\itshape \LaTeX\ User's Guide.}

Frequently-used parameters, or combinations of parameters, include:
\begin{itemize}
\item {\verb|anonymous,review|}: Suitable for a ``dual-anonymous''
  conference submission. Anonymizes the work and includes line
  numbers. Use with the \verb|\acmSubmissionID| command to print the
  submission's unique ID on each page of the work.
\item{\verb|authorversion|}: Produces a version of the work suitable
  for posting by the author.
\item{\verb|screen|}: Produces colored hyperlinks.
\end{itemize}

This document uses the following string as the first command in the
source file:
\begin{verbatim}
\documentclass[sigconf]{acmart}
\end{verbatim}

\section{Modifications}

Modifying the template --- including but not limited to: adjusting
margins, typeface sizes, line spacing, paragraph and list definitions,
and the use of the \verb|\vspace| command to manually adjust the
vertical spacing between elements of your work --- is not allowed.

{\bfseries Your document will be returned to you for revision if
  modifications are discovered.}

\section{Typefaces}

The ``\verb|acmart|'' document class requires the use of the
``Libertine'' typeface family. Your \TeX\ installation should include
this set of packages. Please do not substitute other typefaces. The
``\verb|lmodern|'' and ``\verb|ltimes|'' packages should not be used,
as they will override the built-in typeface families.

\section{Title Information}

The title of your work should use capital letters appropriately -
\url{https://capitalizemytitle.com/} has useful rules for
capitalization. Use the {\verb|title|} command to define the title of
your work. If your work has a subtitle, define it with the
{\verb|subtitle|} command.  Do not insert line breaks in your title.

If your title is lengthy, you must define a short version to be used
in the page headers, to prevent overlapping text. The \verb|title|
command has a ``short title'' parameter:
\begin{verbatim}
  \title[short title]{full title}
\end{verbatim}

\section{Authors and Affiliations}

Each author must be defined separately for accurate metadata
identification. Multiple authors may share one affiliation. Authors'
names should not be abbreviated; use full first names wherever
possible. Include authors' e-mail addresses whenever possible.

Grouping authors' names or e-mail addresses, or providing an ``e-mail
alias,'' as shown below, is not acceptable:
\begin{verbatim}
  \author{Brooke Aster, David Mehldau}
  \email{dave,judy,steve@university.edu}
  \email{firstname.lastname@phillips.org}
\end{verbatim}

The \verb|authornote| and \verb|authornotemark| commands allow a note
to apply to multiple authors --- for example, if the first two authors
of an article contributed equally to the work.

If your author list is lengthy, you must define a shortened version of
the list of authors to be used in the page headers, to prevent
overlapping text. The following command should be placed just after
the last \verb|\author{}| definition:
\begin{verbatim}
  \renewcommand{\shortauthors}{McCartney, et al.}
\end{verbatim}
Omitting this command will force the use of a concatenated list of all
of the authors' names, which may result in overlapping text in the
page headers.

The article template's documentation, available at
\url{https://www.acm.org/publications/proceedings-template}, has a
complete explanation of these commands and tips for their effective
use.

Note that authors' addresses are mandatory for journal articles.

\section{Rights Information}

Authors of any work published by ACM will need to complete a rights
form. Depending on the kind of work, and the rights management choice
made by the author, this may be copyright transfer, permission,
license, or an OA (open access) agreement.

Regardless of the rights management choice, the author will receive a
copy of the completed rights form once it has been submitted. This
form contains \LaTeX\ commands that must be copied into the source
document. When the document source is compiled, these commands and
their parameters add formatted text to several areas of the final
document:
\begin{itemize}
\item the ``ACM Reference Format'' text on the first page.
\item the ``rights management'' text on the first page.
\item the conference information in the page header(s).
\end{itemize}

Rights information is unique to the work; if you are preparing several
works for an event, make sure to use the correct set of commands with
each of the works.

The ACM Reference Format text is required for all articles over one
page in length, and is optional for one-page articles (abstracts).

\section{CCS Concepts and User-Defined Keywords}

Two elements of the ``acmart'' document class provide powerful
taxonomic tools for you to help readers find your work in an online
search.

The ACM Computing Classification System ---
\url{https://www.acm.org/publications/class-2012} --- is a set of
classifiers and concepts that describe the computing
discipline. Authors can select entries from this classification
system, via \url{https://dl.acm.org/ccs/ccs.cfm}, and generate the
commands to be included in the \LaTeX\ source.

User-defined keywords are a comma-separated list of words and phrases
of the authors' choosing, providing a more flexible way of describing
the research being presented.

CCS concepts and user-defined keywords are required for for all
articles over two pages in length, and are optional for one- and
two-page articles (or abstracts).

\section{Sectioning Commands}

Your work should use standard \LaTeX\ sectioning commands:
\verb|section|, \verb|subsection|, \verb|subsubsection|, and
\verb|paragraph|. They should be numbered; do not remove the numbering
from the commands.

Simulating a sectioning command by setting the first word or words of
a paragraph in boldface or italicized text is {\bfseries not allowed.}

\section{Tables}

The ``\verb|acmart|'' document class includes the ``\verb|booktabs|''
package --- \url{https://ctan.org/pkg/booktabs} --- for preparing
high-quality tables.

Table captions are placed {\itshape above} the table.

Because tables cannot be split across pages, the best placement for
them is typically the top of the page nearest their initial cite.  To
ensure this proper ``floating'' placement of tables, use the
environment \textbf{table} to enclose the table's contents and the
table caption.  The contents of the table itself must go in the
\textbf{tabular} environment, to be aligned properly in rows and
columns, with the desired horizontal and vertical rules.  Again,
detailed instructions on \textbf{tabular} material are found in the
\textit{\LaTeX\ User's Guide}.

Immediately following this sentence is the point at which
Table~\ref{tab:freq} is included in the input file; compare the
placement of the table here with the table in the printed output of
this document.

\begin{table}
  \caption{Frequency of Special Characters}
  \label{tab:freq}
  \begin{tabular}{ccl}
    \toprule
    Non-English or Math&Frequency&Comments\\
    \midrule
    \O & 1 in 1,000& For Swedish names\\
    $\pi$ & 1 in 5& Common in math\\
    \$ & 4 in 5 & Used in business\\
    $\Psi^2_1$ & 1 in 40,000& Unexplained usage\\
  \bottomrule
\end{tabular}
\end{table}

To set a wider table, which takes up the whole width of the page's
live area, use the environment \textbf{table*} to enclose the table's
contents and the table caption.  As with a single-column table, this
wide table will ``float'' to a location deemed more
desirable. Immediately following this sentence is the point at which
Table~\ref{tab:commands} is included in the input file; again, it is
instructive to compare the placement of the table here with the table
in the printed output of this document.

\begin{table*}
  \caption{Some Typical Commands}
  \label{tab:commands}
  \begin{tabular}{ccl}
    \toprule
    Command &A Number & Comments\\
    \midrule
    \texttt{{\char'134}author} & 100& Author \\
    \texttt{{\char'134}table}& 300 & For tables\\
    \texttt{{\char'134}table*}& 400& For wider tables\\
    \bottomrule
  \end{tabular}
\end{table*}

Always use midrule to separate table header rows from data rows, and
use it only for this purpose. This enables assistive technologies to
recognise table headers and support their users in navigating tables
more easily.

\section{Math Equations}
You may want to display math equations in three distinct styles:
inline, numbered or non-numbered display.  Each of the three are
discussed in the next sections.

\subsection{Inline (In-text) Equations}
A formula that appears in the running text is called an inline or
in-text formula.  It is produced by the \textbf{math} environment,
which can be invoked with the usual
\texttt{{\char'134}begin\,\ldots{\char'134}end} construction or with
the short form \texttt{\$\,\ldots\$}. You can use any of the symbols
and structures, from $\alpha$ to $\omega$, available in
\LaTeX~\cite{Lamport:LaTeX}; this section will simply show a few
examples of in-text equations in context. Notice how this equation:
\begin{math}
  \lim_{n\rightarrow \infty}x=0
\end{math},
set here in in-line math style, looks slightly different when
set in display style.  (See next section).

\subsection{Display Equations}
A numbered display equation---one set off by vertical space from the
text and centered horizontally---is produced by the \textbf{equation}
environment. An unnumbered display equation is produced by the
\textbf{displaymath} environment.

Again, in either environment, you can use any of the symbols and
structures available in \LaTeX\@; this section will just give a couple
of examples of display equations in context.  First, consider the
equation, shown as an inline equation above:
\begin{equation}
  \lim_{n\rightarrow \infty}x=0
\end{equation}
Notice how it is formatted somewhat differently in
the \textbf{displaymath}
environment.  Now, we'll enter an unnumbered equation:
\begin{displaymath}
  \sum_{i=0}^{\infty} x + 1
\end{displaymath}
and follow it with another numbered equation:
\begin{equation}
  \sum_{i=0}^{\infty}x_i=\int_{0}^{\pi+2} f
\end{equation}
just to demonstrate \LaTeX's able handling of numbering.

\section{Figures}

The ``\verb|figure|'' environment should be used for figures. One or
more images can be placed within a figure. If your figure contains
third-party material, you must clearly identify it as such, as shown
in the example below.
\begin{figure}[h]
  \centering
  \includegraphics[width=\linewidth]{sample-franklin}
  \caption{1907 Franklin Model D roadster. Photograph by Harris \&
    Ewing, Inc. [Public domain], via Wikimedia
    Commons. (\url{https://goo.gl/VLCRBB}).}
  \Description{A woman and a girl in white dresses sit in an open car.}
\end{figure}

Your figures should contain a caption which describes the figure to
the reader.

Figure captions are placed {\itshape below} the figure.

Every figure should also have a figure description unless it is purely
decorative. These descriptions convey what’s in the image to someone
who cannot see it. They are also used by search engine crawlers for
indexing images, and when images cannot be loaded.

A figure description must be unformatted plain text less than 2000
characters long (including spaces).  {\bfseries Figure descriptions
  should not repeat the figure caption – their purpose is to capture
  important information that is not already provided in the caption or
  the main text of the paper.} For figures that convey important and
complex new information, a short text description may not be
adequate. More complex alternative descriptions can be placed in an
appendix and referenced in a short figure description. For example,
provide a data table capturing the information in a bar chart, or a
structured list representing a graph.  For additional information
regarding how best to write figure descriptions and why doing this is
so important, please see
\url{https://www.acm.org/publications/taps/describing-figures/}.

\subsection{The ``Teaser Figure''}

A ``teaser figure'' is an image, or set of images in one figure, that
are placed after all author and affiliation information, and before
the body of the article, spanning the page. If you wish to have such a
figure in your article, place the command immediately before the
\verb|\maketitle| command:
\begin{verbatim}
  \begin{teaserfigure}
    \includegraphics[width=\textwidth]{sampleteaser}
    \caption{figure caption}
    \Description{figure description}
  \end{teaserfigure}
\end{verbatim}

\section{Citations and Bibliographies}

The use of \BibTeX\ for the preparation and formatting of one's
references is strongly recommended. Authors' names should be complete
--- use full first names (``Donald E. Knuth'') not initials
(``D. E. Knuth'') --- and the salient identifying features of a
reference should be included: title, year, volume, number, pages,
article DOI, etc.

The bibliography is included in your source document with these two
commands, placed just before the \verb|\end{document}| command:
\begin{verbatim}
  \bibliographystyle{ACM-Reference-Format}
  \bibliography{bibfile}
\end{verbatim}
where ``\verb|bibfile|'' is the name, without the ``\verb|.bib|''
suffix, of the \BibTeX\ file.

Citations and references are numbered by default. A small number of
ACM publications have citations and references formatted in the
``author year'' style; for these exceptions, please include this
command in the {\bfseries preamble} (before the command
``\verb|\begin{document}|'') of your \LaTeX\ source:
\begin{verbatim}
  \citestyle{acmauthoryear}
\end{verbatim}

  Some examples.  A paginated journal article \cite{Abril07}, an
  enumerated journal article \cite{Cohen07}, a reference to an entire
  issue \cite{JCohen96}, a monograph (whole book) \cite{Kosiur01}, a
  monograph/whole book in a series (see 2a in spec. document)
  \cite{Harel79}, a divisible-book such as an anthology or compilation
  \cite{Editor00} followed by the same example, however we only output
  the series if the volume number is given \cite{Editor00a} (so
  Editor00a's series should NOT be present since it has no vol. no.),
  a chapter in a divisible book \cite{Spector90}, a chapter in a
  divisible book in a series \cite{Douglass98}, a multi-volume work as
  book \cite{Knuth97}, a couple of articles in a proceedings (of a
  conference, symposium, workshop for example) (paginated proceedings
  article) \cite{Andler79, Hagerup1993}, a proceedings article with
  all possible elements \cite{Smith10}, an example of an enumerated
  proceedings article \cite{VanGundy07}, an informally published work
  \cite{Harel78}, a couple of preprints \cite{Bornmann2019,
    AnzarootPBM14}, a doctoral dissertation \cite{Clarkson85}, a
  master's thesis: \cite{anisi03}, an online document / world wide web
  resource \cite{Thornburg01, Ablamowicz07, Poker06}, a video game
  (Case 1) \cite{Obama08} and (Case 2) \cite{Novak03} and \cite{Lee05}
  and (Case 3) a patent \cite{JoeScientist001}, work accepted for
  publication \cite{rous08}, 'YYYYb'-test for prolific author
  \cite{SaeediMEJ10} and \cite{SaeediJETC10}. Other cites might
  contain 'duplicate' DOI and URLs (some SIAM articles)
  \cite{Kirschmer:2010:AEI:1958016.1958018}. Boris / Barbara Beeton:
  multi-volume works as books \cite{MR781536} and \cite{MR781537}. A
  couple of citations with DOIs:
  \cite{2004:ITE:1009386.1010128,Kirschmer:2010:AEI:1958016.1958018}. Online
  citations: \cite{TUGInstmem, Thornburg01, CTANacmart}. Artifacts:
  \cite{R} and \cite{UMassCitations}.

\section{Acknowledgments}

Identification of funding sources and other support, and thanks to
individuals and groups that assisted in the research and the
preparation of the work should be included in an acknowledgment
section, which is placed just before the reference section in your
document.

This section has a special environment:
\begin{verbatim}
  \begin{acks}
  ...
  \end{acks}
\end{verbatim}
so that the information contained therein can be more easily collected
during the article metadata extraction phase, and to ensure
consistency in the spelling of the section heading.

Authors should not prepare this section as a numbered or unnumbered {\verb|\section|}; please use the ``{\verb|acks|}'' environment.

\section{Appendices}

If your work needs an appendix, add it before the
``\verb|\end{document}|'' command at the conclusion of your source
document.

Start the appendix with the ``\verb|appendix|'' command:
\begin{verbatim}
  \appendix
\end{verbatim}
and note that in the appendix, sections are lettered, not
numbered. This document has two appendices, demonstrating the section
and subsection identification method.

\section{Multi-language papers}

Papers may be written in languages other than English or include
titles, subtitles, keywords and abstracts in different languages (as a
rule, a paper in a language other than English should include an
English title and an English abstract).  Use \verb|language=...| for
every language used in the paper.  The last language indicated is the
main language of the paper.  For example, a French paper with
additional titles and abstracts in English and German may start with
the following command
\begin{verbatim}
\documentclass[sigconf, language=english, language=german,
               language=french]{acmart}
\end{verbatim}

The title, subtitle, keywords and abstract will be typeset in the main
language of the paper.  The commands \verb|\translatedXXX|, \verb|XXX|
begin title, subtitle and keywords, can be used to set these elements
in the other languages.  The environment \verb|translatedabstract| is
used to set the translation of the abstract.  These commands and
environment have a mandatory first argument: the language of the
second argument.  See \verb|sample-sigconf-i13n.tex| file for examples
of their usage.

\section{SIGCHI Extended Abstracts}

The ``\verb|sigchi-a|'' template style (available only in \LaTeX\ and
not in Word) produces a landscape-orientation formatted article, with
a wide left margin. Three environments are available for use with the
``\verb|sigchi-a|'' template style, and produce formatted output in
the margin:
\begin{itemize}
\item {\verb|sidebar|}:  Place formatted text in the margin.
\item {\verb|marginfigure|}: Place a figure in the margin.
\item {\verb|margintable|}: Place a table in the margin.
\end{itemize}
\fi


\begin{acks}
Soumyadeep Roy is supported by the Institute Ph.D. Fellowship at the Indian Institute of Technology Kharagpur. This research was partially funded by the Federal Ministry of Education and Research (BMBF), Germany under the project LeibnizKILabor with grant No. 01DD20003.
\end{acks}

\bibliographystyle{ACM-Reference-Format}
\bibliography{sample-base}


\begin{thebibliography}{50}

%%% the Web address.

\ifx \showCODEN    \undefined \def \showCODEN     #1{\unskip}     \fi
\ifx \showDOI      \undefined \def \showDOI       #1{#1}\fi
\ifx \showISBNx    \undefined \def \showISBNx     #1{\unskip}     \fi
\ifx \showISBNxiii \undefined \def \showISBNxiii  #1{\unskip}     \fi
\ifx \showISSN     \undefined \def \showISSN      #1{\unskip}     \fi
\ifx \showLCCN     \undefined \def \showLCCN      #1{\unskip}     \fi
\ifx \shownote     \undefined \def \shownote      #1{#1}          \fi
\ifx \showarticletitle \undefined \def \showarticletitle #1{#1}   \fi
\ifx \showURL      \undefined \def \showURL       {\relax}        \fi
% invisible to TeX
\providecommand\bibfield[2]{#2}
\providecommand\bibinfo[2]{#2}
\providecommand\natexlab[1]{#1}
\providecommand\showeprint[2][]{arXiv:#2}

\bibitem[Adams et~al\mbox{.}(2023)]%
        {adams2023meta}
\bibfield{author}{\bibinfo{person}{Griffin Adams}, \bibinfo{person}{Jason Zuckerg}, {and} \bibinfo{person}{No{\'e}mie Elhadad}.} \bibinfo{year}{2023}\natexlab{}.
\newblock \showarticletitle{A meta-evaluation of faithfulness metrics for long-form hospital-course summarization}. In \bibinfo{booktitle}{\emph{Machine Learning for Healthcare Conference}}. PMLR, \bibinfo{pages}{2--30}.
\newblock


\bibitem[Besta et~al\mbox{.}(2024)]%
        {besta2023graph}
\bibfield{author}{\bibinfo{person}{Maciej Besta}, \bibinfo{person}{Nils Blach}, \bibinfo{person}{Ales Kubicek}, {et~al\mbox{.}}} \bibinfo{year}{2024}\natexlab{}.
\newblock \showarticletitle{Graph of Thoughts: Solving Elaborate Problems with Large Language Models}.
\newblock \bibinfo{journal}{\emph{Proceedings of the AAAI Conference on Artificial Intelligence}} \bibinfo{volume}{38}, \bibinfo{number}{16} (\bibinfo{date}{Mar.} \bibinfo{year}{2024}), \bibinfo{pages}{17682--17690}.
\newblock
\urldef\tempurl%
\url{https://doi.org/10.1609/aaai.v38i16.29720}
\showDOI{\tempurl}


\bibitem[Chen et~al\mbox{.}(2023a)]%
        {chenarjun2023}
\bibfield{author}{\bibinfo{person}{Anjun Chen}, \bibinfo{person}{Drake~O Chen}, {and} \bibinfo{person}{Lu Tian}.} \bibinfo{year}{2023}\natexlab{a}.
\newblock \showarticletitle{{Benchmarking the symptom-checking capabilities of ChatGPT for a broad range of diseases}}.
\newblock \bibinfo{journal}{\emph{Journal of the American Medical Informatics Association}} (\bibinfo{date}{12} \bibinfo{year}{2023}), \bibinfo{pages}{ocad245}.
\newblock
\showISSN{1527-974X}
\urldef\tempurl%
\url{https://doi.org/10.1093/jamia/ocad245}
\showDOI{\tempurl}


\bibitem[Chen et~al\mbox{.}(2023b)]%
        {chen2023chatgpt}
\bibfield{author}{\bibinfo{person}{Lingjiao Chen}, \bibinfo{person}{Matei Zaharia}, {and} \bibinfo{person}{James Zou}.} \bibinfo{year}{2023}\natexlab{b}.
\newblock \showarticletitle{How is ChatGPT's behavior changing over time?}
\newblock \bibinfo{journal}{\emph{arXiv preprint arXiv:2307.09009}} (\bibinfo{year}{2023}).
\newblock


\bibitem[Chen et~al\mbox{.}(2021)]%
        {chen2021codex}
\bibfield{author}{\bibinfo{person}{Mark Chen}, \bibinfo{person}{Jerry Tworek}, \bibinfo{person}{Heewoo Jun}, {et~al\mbox{.}}} \bibinfo{year}{2021}\natexlab{}.
\newblock \bibinfo{title}{Evaluating Large Language Models Trained on Code}.
\newblock
\newblock
\showeprint[arxiv]{2107.03374}~[cs.LG]


\bibitem[Chern et~al\mbox{.}(2023)]%
        {chern2023factool}
\bibfield{author}{\bibinfo{person}{I Chern}, \bibinfo{person}{Steffi Chern}, \bibinfo{person}{Shiqi Chen}, {et~al\mbox{.}}} \bibinfo{year}{2023}\natexlab{}.
\newblock \showarticletitle{FacTool: Factuality Detection in Generative AI--A Tool Augmented Framework for Multi-Task and Multi-Domain Scenarios}.
\newblock \bibinfo{journal}{\emph{arXiv preprint arXiv:2307.13528}} (\bibinfo{year}{2023}).
\newblock


\bibitem[Dash et~al\mbox{.}(2023)]%
        {dash2023evaluation}
\bibfield{author}{\bibinfo{person}{Debadutta Dash}, \bibinfo{person}{Rahul Thapa}, \bibinfo{person}{Juan~M Banda}, {et~al\mbox{.}}} \bibinfo{year}{2023}\natexlab{}.
\newblock \showarticletitle{Evaluation of GPT-3.5 and GPT-4 for supporting real-world information needs in healthcare delivery}.
\newblock \bibinfo{journal}{\emph{arXiv preprint arXiv:2304.13714}} (\bibinfo{year}{2023}).
\newblock


\bibitem[Fleisig et~al\mbox{.}(2023)]%
        {fleisig2023majority}
\bibfield{author}{\bibinfo{person}{Eve Fleisig}, \bibinfo{person}{Rediet Abebe}, {and} \bibinfo{person}{Dan Klein}.} \bibinfo{year}{2023}\natexlab{}.
\newblock \showarticletitle{When the Majority is Wrong: Modeling Annotator Disagreement for Subjective Tasks}. In \bibinfo{booktitle}{\emph{Proceedings of the 2023 Conference on Empirical Methods in Natural Language Processing}}. \bibinfo{pages}{6715--6726}.
\newblock


\bibitem[Han et~al\mbox{.}(2023)]%
        {han2023medalpaca}
\bibfield{author}{\bibinfo{person}{Tianyu Han}, \bibinfo{person}{Lisa~C. Adams}, \bibinfo{person}{Jens-Michalis Papaioannou}, {et~al\mbox{.}}} \bibinfo{year}{2023}\natexlab{}.
\newblock \bibinfo{title}{MedAlpaca -- An Open-Source Collection of Medical Conversational AI Models and Training Data}.
\newblock
\newblock
\showeprint[arxiv]{2304.08247}~[cs.CL]


\bibitem[Harrigian et~al\mbox{.}(2023)]%
        {harrigian2023characterization}
\bibfield{author}{\bibinfo{person}{Keith Harrigian}, \bibinfo{person}{Ayah Zirikly}, \bibinfo{person}{Brant Chee}, {et~al\mbox{.}}} \bibinfo{year}{2023}\natexlab{}.
\newblock \showarticletitle{Characterization of stigmatizing language in medical records}. In \bibinfo{booktitle}{\emph{Proceedings of the 61st Annual Meeting of the Association for Computational Linguistics (Volume 2: Short Papers)}}. \bibinfo{pages}{312--329}.
\newblock


\bibitem[Hendrycks et~al\mbox{.}(2021)]%
        {hendrycks2021measuring}
\bibfield{author}{\bibinfo{person}{Dan Hendrycks}, \bibinfo{person}{Collin Burns}, \bibinfo{person}{Steven Basart}, \bibinfo{person}{Andy Zou}, \bibinfo{person}{Mantas Mazeika}, \bibinfo{person}{Dawn Song}, {and} \bibinfo{person}{Jacob Steinhardt}.} \bibinfo{year}{2021}\natexlab{}.
\newblock \showarticletitle{Measuring Massive Multitask Language Understanding}. In \bibinfo{booktitle}{\emph{International Conference on Learning Representations}}.
\newblock
\urldef\tempurl%
\url{https://openreview.net/forum?id=d7KBjmI3GmQ}
\showURL{%
\tempurl}


\bibitem[Huang et~al\mbox{.}(2023)]%
        {huang2023survey}
\bibfield{author}{\bibinfo{person}{Lei Huang}, \bibinfo{person}{Weijiang Yu}, \bibinfo{person}{Weitao Ma}, {et~al\mbox{.}}} \bibinfo{year}{2023}\natexlab{}.
\newblock \showarticletitle{A survey on hallucination in large language models: Principles, taxonomy, challenges, and open questions}.
\newblock \bibinfo{journal}{\emph{arXiv preprint arXiv:2311.05232}} (\bibinfo{year}{2023}).
\newblock


\bibitem[Jin et~al\mbox{.}(2021)]%
        {jin2021disease}
\bibfield{author}{\bibinfo{person}{Di Jin}, \bibinfo{person}{Eileen Pan}, \bibinfo{person}{Nassim Oufattole}, {et~al\mbox{.}}} \bibinfo{year}{2021}\natexlab{}.
\newblock \showarticletitle{What disease does this patient have? a large-scale open domain question answering dataset from medical exams}.
\newblock \bibinfo{journal}{\emph{Applied Sciences}} \bibinfo{volume}{11}, \bibinfo{number}{14} (\bibinfo{year}{2021}), \bibinfo{pages}{6421}.
\newblock


\bibitem[Jin et~al\mbox{.}(2024)]%
        {jin2024hidden}
\bibfield{author}{\bibinfo{person}{Qiao Jin}, \bibinfo{person}{Fangyuan Chen}, \bibinfo{person}{Yiliang Zhou}, {et~al\mbox{.}}} \bibinfo{year}{2024}\natexlab{}.
\newblock \showarticletitle{Hidden flaws behind expert-level accuracy of gpt-4 vision in medicine}.
\newblock \bibinfo{journal}{\emph{arXiv preprint arXiv:2401.08396}} (\bibinfo{year}{2024}).
\newblock


\bibitem[Jin et~al\mbox{.}(2019)]%
        {jin-etal-2019-pubmedqa}
\bibfield{author}{\bibinfo{person}{Qiao Jin}, \bibinfo{person}{Bhuwan Dhingra}, \bibinfo{person}{Zhengping Liu}, \bibinfo{person}{William Cohen}, {and} \bibinfo{person}{Xinghua Lu}.} \bibinfo{year}{2019}\natexlab{}.
\newblock \showarticletitle{{P}ub{M}ed{QA}: A Dataset for Biomedical Research Question Answering}. In \bibinfo{booktitle}{\emph{Proceedings of the 2019 Conference on Empirical Methods in Natural Language Processing and the 9th International Joint Conference on Natural Language Processing (EMNLP-IJCNLP)}}. \bibinfo{address}{Hong Kong, China}, \bibinfo{pages}{2567--2577}.
\newblock
\urldef\tempurl%
\url{https://doi.org/10.18653/v1/D19-1259}
\showDOI{\tempurl}


\bibitem[Johnson et~al\mbox{.}(2019)]%
        {johnson2019mimic}
\bibfield{author}{\bibinfo{person}{Alistair~EW Johnson}, \bibinfo{person}{Tom~J Pollard}, \bibinfo{person}{Seth~J Berkowitz}, {et~al\mbox{.}}} \bibinfo{year}{2019}\natexlab{}.
\newblock \showarticletitle{MIMIC-CXR, a de-identified publicly available database of chest radiographs with free-text reports}.
\newblock \bibinfo{journal}{\emph{Scientific data}} \bibinfo{volume}{6}, \bibinfo{number}{1} (\bibinfo{year}{2019}), \bibinfo{pages}{317}.
\newblock


\bibitem[Kabir et~al\mbox{.}(2023)]%
        {kabir2023answers}
\bibfield{author}{\bibinfo{person}{Samia Kabir}, \bibinfo{person}{David~N Udo-Imeh}, \bibinfo{person}{Bonan Kou}, {and} \bibinfo{person}{Tianyi Zhang}.} \bibinfo{year}{2023}\natexlab{}.
\newblock \showarticletitle{Who answers it better? an in-depth analysis of ChatGPT and stack overflow answers to software engineering questions}.
\newblock \bibinfo{journal}{\emph{arXiv preprint arXiv:2308.02312}} (\bibinfo{year}{2023}).
\newblock


\bibitem[Kanjee et~al\mbox{.}(2023)]%
        {kanjee2023}
\bibfield{author}{\bibinfo{person}{Zahir Kanjee}, \bibinfo{person}{Byron Crowe}, {and} \bibinfo{person}{Adam Rodman}.} \bibinfo{year}{2023}\natexlab{}.
\newblock \showarticletitle{{Accuracy of a Generative Artificial Intelligence Model in a Complex Diagnostic Challenge}}.
\newblock \bibinfo{journal}{\emph{JAMA}} \bibinfo{volume}{330}, \bibinfo{number}{1} (\bibinfo{date}{07} \bibinfo{year}{2023}), \bibinfo{pages}{78--80}.
\newblock
\showISSN{0098-7484}
\urldef\tempurl%
\url{https://doi.org/10.1001/jama.2023.8288}
\showDOI{\tempurl}


\bibitem[Kilicoglu et~al\mbox{.}(2020)]%
        {Kilicoglu2020}
\bibfield{author}{\bibinfo{person}{Halil Kilicoglu}, \bibinfo{person}{Graciela Rosemblat}, \bibinfo{person}{Marcelo Fiszman}, {and} \bibinfo{person}{Dongwook Shin}.} \bibinfo{year}{2020}\natexlab{}.
\newblock \showarticletitle{Broad-coverage biomedical relation extraction with SemRep}.
\newblock \bibinfo{journal}{\emph{BMC Bioinformatics}} \bibinfo{volume}{21}, \bibinfo{number}{1} (\bibinfo{date}{14 May} \bibinfo{year}{2020}), \bibinfo{pages}{188}.
\newblock
\showISSN{1471-2105}
\urldef\tempurl%
\url{https://doi.org/10.1186/s12859-020-3517-7}
\showDOI{\tempurl}


\bibitem[Li et~al\mbox{.}(2021)]%
        {li2021construction}
\bibfield{author}{\bibinfo{person}{Zepeng Li}, \bibinfo{person}{Yufeng Zhang}, \bibinfo{person}{Rikui Huang}, {et~al\mbox{.}}} \bibinfo{year}{2021}\natexlab{}.
\newblock \showarticletitle{Construction of Depression Knowledge Graph Based on Biomedical Literature}. In \bibinfo{booktitle}{\emph{2021 IEEE International Conference on Bioinformatics and Biomedicine (BIBM)}}. IEEE, \bibinfo{pages}{1849--1855}.
\newblock


\bibitem[Liévin et~al\mbox{.}(2024)]%
        {liévin2023large}
\bibfield{author}{\bibinfo{person}{Valentin Liévin}, \bibinfo{person}{Christoffer~Egeberg Hother}, \bibinfo{person}{Andreas~Geert Motzfeldt}, {and} \bibinfo{person}{Ole Winther}.} \bibinfo{year}{2024}\natexlab{}.
\newblock \showarticletitle{Can large language models reason about medical questions?}
\newblock \bibinfo{journal}{\emph{Patterns}} \bibinfo{volume}{5}, \bibinfo{number}{3} (\bibinfo{year}{2024}), \bibinfo{pages}{100943}.
\newblock
\showISSN{2666-3899}
\urldef\tempurl%
\url{https://doi.org/10.1016/j.patter.2024.100943}
\showDOI{\tempurl}


\bibitem[McDuff et~al\mbox{.}(2023)]%
        {mcduff2023accurate}
\bibfield{author}{\bibinfo{person}{Daniel McDuff}, \bibinfo{person}{Mike Schaekermann}, \bibinfo{person}{Tao Tu}, {et~al\mbox{.}}} \bibinfo{year}{2023}\natexlab{}.
\newblock \bibinfo{title}{Towards Accurate Differential Diagnosis with Large Language Models}.
\newblock
\newblock
\showeprint[arxiv]{2312.00164}~[cs.CY]


\bibitem[Moramarco et~al\mbox{.}(2022)]%
        {moramarco2022human}
\bibfield{author}{\bibinfo{person}{Francesco Moramarco}, \bibinfo{person}{{Alex Papadopoulos} Korfiatis}, \bibinfo{person}{Mark Perera}, {et~al\mbox{.}}} \bibinfo{year}{2022}\natexlab{}.
\newblock \showarticletitle{Human Evaluation and Correlation with Automatic Metrics in Consultation Note Generation}. In \bibinfo{booktitle}{\emph{Proceedings of the 60th Annual Meeting of the Association for Computational Linguistics (Volume 1: Long Papers)}}, Vol.~\bibinfo{volume}{1}. \bibinfo{pages}{5739–5754}.
\newblock
\urldef\tempurl%
\url{https://doi.org/10.48550/ARXIV.2204.00447}
\showDOI{\tempurl}


\bibitem[Nori et~al\mbox{.}(2023)]%
        {nori2023capabilities}
\bibfield{author}{\bibinfo{person}{Harsha Nori}, \bibinfo{person}{Nicholas King}, \bibinfo{person}{Scott~Mayer McKinney}, \bibinfo{person}{Dean Carignan}, {and} \bibinfo{person}{Eric Horvitz}.} \bibinfo{year}{2023}\natexlab{}.
\newblock \showarticletitle{Capabilities of gpt-4 on medical challenge problems}.
\newblock \bibinfo{journal}{\emph{arXiv preprint arXiv:2303.13375}} (\bibinfo{year}{2023}).
\newblock


\bibitem[OpenAI et~al\mbox{.}(2023)]%
        {openai2023gpt4}
\bibfield{author}{\bibinfo{person}{OpenAI}, \bibinfo{person}{:}, \bibinfo{person}{Josh Achiam}, \bibinfo{person}{Steven Adler}, {et~al\mbox{.}}} \bibinfo{year}{2023}\natexlab{}.
\newblock \bibinfo{title}{GPT-4 Technical Report}.
\newblock
\newblock
\showeprint[arxiv]{2303.08774}~[cs.CL]


\bibitem[Pal et~al\mbox{.}(2022)]%
        {pmlr-v174-pal22a}
\bibfield{author}{\bibinfo{person}{Ankit Pal}, \bibinfo{person}{Logesh~Kumar Umapathi}, {and} \bibinfo{person}{Malaikannan Sankarasubbu}.} \bibinfo{year}{2022}\natexlab{}.
\newblock \showarticletitle{MedMCQA: A Large-scale Multi-Subject Multi-Choice Dataset for Medical domain Question Answering}. In \bibinfo{booktitle}{\emph{Proceedings of the Conference on Health, Inference, and Learning}} \emph{(\bibinfo{series}{Proceedings of Machine Learning Research}, Vol.~\bibinfo{volume}{174})}. \bibinfo{pages}{248--260}.
\newblock
\urldef\tempurl%
\url{https://proceedings.mlr.press/v174/pal22a.html}
\showURL{%
\tempurl}


\bibitem[Pei et~al\mbox{.}(2022)]%
        {pei-etal-2022-potato}
\bibfield{author}{\bibinfo{person}{Jiaxin Pei}, \bibinfo{person}{Aparna Ananthasubramaniam}, \bibinfo{person}{Xingyao Wang}, {et~al\mbox{.}}} \bibinfo{year}{2022}\natexlab{}.
\newblock \showarticletitle{{POTATO}: The Portable Text Annotation Tool}. In \bibinfo{booktitle}{\emph{Proceedings of the 2022 Conference on Empirical Methods in Natural Language Processing: System Demonstrations}}. \bibinfo{address}{Abu Dhabi, UAE}, \bibinfo{pages}{327--337}.
\newblock
\urldef\tempurl%
\url{https://doi.org/10.18653/v1/2022.emnlp-demos.33}
\showDOI{\tempurl}


\bibitem[Pei and Jurgens(2023)]%
        {pei2023annotator}
\bibfield{author}{\bibinfo{person}{Jiaxin Pei} {and} \bibinfo{person}{David Jurgens}.} \bibinfo{year}{2023}\natexlab{}.
\newblock \showarticletitle{When Do Annotator Demographics Matter? Measuring the Influence of Annotator Demographics with the POPQUORN Dataset}.
\newblock \bibinfo{journal}{\emph{arXiv preprint arXiv:2306.06826}} (\bibinfo{year}{2023}).
\newblock


\bibitem[Peng et~al\mbox{.}(2023)]%
        {Peng2023gatortron}
\bibfield{author}{\bibinfo{person}{Cheng Peng}, \bibinfo{person}{Xi Yang}, \bibinfo{person}{Aokun Chen}, {et~al\mbox{.}}} \bibinfo{year}{2023}\natexlab{}.
\newblock \showarticletitle{A study of generative large language model for medical research and healthcare}.
\newblock \bibinfo{journal}{\emph{npj Digital Medicine}} \bibinfo{volume}{6}, \bibinfo{number}{1} (\bibinfo{date}{16 Nov} \bibinfo{year}{2023}), \bibinfo{pages}{210}.
\newblock
\showISSN{2398-6352}
\urldef\tempurl%
\url{https://doi.org/10.1038/s41746-023-00958-w}
\showDOI{\tempurl}


\bibitem[Poddar et~al\mbox{.}(2022)]%
        {sohamcaves2022}
\bibfield{author}{\bibinfo{person}{Soham Poddar}, \bibinfo{person}{Azlaan~Mustafa Samad}, \bibinfo{person}{Rajdeep Mukherjee}, {et~al\mbox{.}}} \bibinfo{year}{2022}\natexlab{}.
\newblock \showarticletitle{CAVES: A Dataset to facilitate Explainable Classification and Summarization of Concerns towards COVID Vaccines}. In \bibinfo{booktitle}{\emph{Proceedings of the 45th International ACM SIGIR Conference on Research and Development in Information Retrieval}} (Madrid, Spain) \emph{(\bibinfo{series}{SIGIR '22})}. \bibinfo{pages}{3154–3164}.
\newblock
\showISBNx{9781450387323}
\urldef\tempurl%
\url{https://doi.org/10.1145/3477495.3531745}
\showDOI{\tempurl}


\bibitem[Prolific(2023)]%
        {prolificwebsite}
Prolific \bibinfo{year}{2023}\natexlab{}.
\newblock
\newblock
\urldef\tempurl%
\url{https://app.prolific.com/}
\showURL{%
\tempurl}


\bibitem[Rindflesch and Fiszman(2003)]%
        {rindflesch2003interaction}
\bibfield{author}{\bibinfo{person}{Thomas~C Rindflesch} {and} \bibinfo{person}{Marcelo Fiszman}.} \bibinfo{year}{2003}\natexlab{}.
\newblock \showarticletitle{The interaction of domain knowledge and linguistic structure in natural language processing: interpreting hypernymic propositions in biomedical text}.
\newblock \bibinfo{journal}{\emph{Journal of biomedical informatics}} \bibinfo{volume}{36}, \bibinfo{number}{6} (\bibinfo{year}{2003}), \bibinfo{pages}{462--477}.
\newblock


\bibitem[Savage et~al\mbox{.}(2024)]%
        {savage2023diagnostic}
\bibfield{author}{\bibinfo{person}{Thomas Savage}, \bibinfo{person}{Ashwin Nayak}, \bibinfo{person}{Robert Gallo}, {et~al\mbox{.}}} \bibinfo{year}{2024}\natexlab{}.
\newblock \showarticletitle{Diagnostic reasoning prompts reveal the potential for large language model interpretability in medicine}.
\newblock \bibinfo{journal}{\emph{npj Digital Medicine}} \bibinfo{volume}{7}, \bibinfo{number}{1} (\bibinfo{date}{24 Jan} \bibinfo{year}{2024}), \bibinfo{pages}{20}.
\newblock
\showISSN{2398-6352}
\urldef\tempurl%
\url{https://doi.org/10.1038/s41746-024-01010-1}
\showDOI{\tempurl}


\bibitem[Schutte et~al\mbox{.}(2021)]%
        {schutte2021discovering}
\bibfield{author}{\bibinfo{person}{Dalton Schutte}, \bibinfo{person}{Jake Vasilakes}, \bibinfo{person}{Anu Bompelli}, {et~al\mbox{.}}} \bibinfo{year}{2021}\natexlab{}.
\newblock \showarticletitle{Discovering novel drug-supplement interactions using a dietary supplements knowledge graph generated from the biomedical literature}.
\newblock \bibinfo{journal}{\emph{arXiv preprint arXiv:2106.12741}} (\bibinfo{year}{2021}).
\newblock


\bibitem[Singhal et~al\mbox{.}(2023a)]%
        {Singhal2023}
\bibfield{author}{\bibinfo{person}{Karan Singhal}, \bibinfo{person}{Shekoofeh Azizi}, \bibinfo{person}{Tao Tu}, {et~al\mbox{.}}} \bibinfo{year}{2023}\natexlab{a}.
\newblock \showarticletitle{Large language models encode clinical knowledge}.
\newblock \bibinfo{journal}{\emph{Nature}} \bibinfo{volume}{620}, \bibinfo{number}{7972} (\bibinfo{date}{01 Aug} \bibinfo{year}{2023}), \bibinfo{pages}{172--180}.
\newblock
\showISSN{1476-4687}
\urldef\tempurl%
\url{https://doi.org/10.1038/s41586-023-06291-2}
\showDOI{\tempurl}


\bibitem[Singhal et~al\mbox{.}(2023b)]%
        {singhal2023medpalm2}
\bibfield{author}{\bibinfo{person}{Karan Singhal}, \bibinfo{person}{Tao Tu}, \bibinfo{person}{Juraj Gottweis}, {et~al\mbox{.}}} \bibinfo{year}{2023}\natexlab{b}.
\newblock \bibinfo{title}{Towards Expert-Level Medical Question Answering with Large Language Models}.
\newblock
\newblock
\showeprint[arxiv]{2305.09617}~[cs.CL]


\bibitem[Taylor et~al\mbox{.}(2022)]%
        {taylor2022galactica}
\bibfield{author}{\bibinfo{person}{Ross Taylor}, \bibinfo{person}{Marcin Kardas}, \bibinfo{person}{Guillem Cucurull}, {et~al\mbox{.}}} \bibinfo{year}{2022}\natexlab{}.
\newblock \bibinfo{title}{Galactica: A Large Language Model for Science}.
\newblock
\newblock
\showeprint[arxiv]{2211.09085}~[cs.CL]


\bibitem[Touvron et~al\mbox{.}(2023)]%
        {touvron2023llama}
\bibfield{author}{\bibinfo{person}{Hugo Touvron}, \bibinfo{person}{Louis Martin}, \bibinfo{person}{Kevin Stone}, {et~al\mbox{.}}} \bibinfo{year}{2023}\natexlab{}.
\newblock \showarticletitle{Llama 2: Open foundation and fine-tuned chat models}.
\newblock \bibinfo{journal}{\emph{arXiv preprint arXiv:2307.09288}} (\bibinfo{year}{2023}).
\newblock


\bibitem[USMLE official website(2024)]%
        {usmlewebsite}
USMLE official website \bibinfo{year}{2024}\natexlab{}.
\newblock
\newblock
\urldef\tempurl%
\url{https://www.usmle.org/about-usmle}
\showURL{%
\tempurl}
\newblock
\shownote{Accessed on February 7, 2024}.


\bibitem[Wadden et~al\mbox{.}(2020)]%
        {wadden2020fact}
\bibfield{author}{\bibinfo{person}{David Wadden}, \bibinfo{person}{Shanchuan Lin}, \bibinfo{person}{Kyle Lo}, {et~al\mbox{.}}} \bibinfo{year}{2020}\natexlab{}.
\newblock \showarticletitle{Fact or Fiction: Verifying Scientific Claims}. In \bibinfo{booktitle}{\emph{Proceedings of the 2020 Conference on Empirical Methods in Natural Language Processing (EMNLP)}}. \bibinfo{address}{Online}, \bibinfo{pages}{7534--7550}.
\newblock
\urldef\tempurl%
\url{https://doi.org/10.18653/v1/2020.emnlp-main.609}
\showDOI{\tempurl}


\bibitem[Wan et~al\mbox{.}(2023)]%
        {wan2023everyone}
\bibfield{author}{\bibinfo{person}{Ruyuan Wan}, \bibinfo{person}{Jaehyung Kim}, {and} \bibinfo{person}{Dongyeop Kang}.} \bibinfo{year}{2023}\natexlab{}.
\newblock \showarticletitle{Everyone’s Voice Matters: Quantifying Annotation Disagreement Using Demographic Information}.
\newblock \bibinfo{journal}{\emph{Proceedings of the AAAI Conference on Artificial Intelligence}} \bibinfo{volume}{37}, \bibinfo{number}{12} (\bibinfo{date}{Jun.} \bibinfo{year}{2023}), \bibinfo{pages}{14523--14530}.
\newblock
\urldef\tempurl%
\url{https://doi.org/10.1609/aaai.v37i12.26698}
\showDOI{\tempurl}


\bibitem[Wei et~al\mbox{.}(2022)]%
        {wei2022chain}
\bibfield{author}{\bibinfo{person}{Jason Wei}, \bibinfo{person}{Xuezhi Wang}, \bibinfo{person}{Dale Schuurmans}, {et~al\mbox{.}}} \bibinfo{year}{2022}\natexlab{}.
\newblock \showarticletitle{Chain-of-thought prompting elicits reasoning in large language models}.
\newblock \bibinfo{journal}{\emph{Advances in Neural Information Processing Systems}}  \bibinfo{volume}{35} (\bibinfo{year}{2022}), \bibinfo{pages}{24824--24837}.
\newblock


\bibitem[Wellbery(2011)]%
        {wellbery2011flaws}
\bibfield{author}{\bibinfo{person}{Caroline Wellbery}.} \bibinfo{year}{2011}\natexlab{}.
\newblock \showarticletitle{Flaws in clinical reasoning: a common cause of diagnostic error}.
\newblock \bibinfo{journal}{\emph{American family physician}} \bibinfo{volume}{84}, \bibinfo{number}{9} (\bibinfo{year}{2011}), \bibinfo{pages}{1042--1048}.
\newblock


\bibitem[Wu et~al\mbox{.}(2023)]%
        {wu2023pmcllama}
\bibfield{author}{\bibinfo{person}{Chaoyi Wu}, \bibinfo{person}{Weixiong Lin}, \bibinfo{person}{Xiaoman Zhang}, \bibinfo{person}{Ya Zhang}, \bibinfo{person}{Yanfeng Wang}, {and} \bibinfo{person}{Weidi Xie}.} \bibinfo{year}{2023}\natexlab{}.
\newblock \bibinfo{title}{PMC-LLaMA: Towards Building Open-source Language Models for Medicine}.
\newblock
\newblock
\showeprint[arxiv]{2304.14454}~[cs.CL]


\bibitem[Xie et~al\mbox{.}(2022)]%
        {xie2022pre}
\bibfield{author}{\bibinfo{person}{Qianqian Xie}, \bibinfo{person}{Jennifer~Amy Bishop}, \bibinfo{person}{Prayag Tiwari}, {and} \bibinfo{person}{Sophia Ananiadou}.} \bibinfo{year}{2022}\natexlab{}.
\newblock \showarticletitle{Pre-trained language models with domain knowledge for biomedical extractive summarization}.
\newblock \bibinfo{journal}{\emph{Knowledge-Based Systems}}  \bibinfo{volume}{252} (\bibinfo{year}{2022}), \bibinfo{pages}{109460}.
\newblock


\bibitem[Xie et~al\mbox{.}(2023a)]%
        {xie2023survey}
\bibfield{author}{\bibinfo{person}{Qianqian Xie}, \bibinfo{person}{Zheheng Luo}, \bibinfo{person}{Benyou Wang}, {and} \bibinfo{person}{Sophia Ananiadou}.} \bibinfo{year}{2023}\natexlab{a}.
\newblock \showarticletitle{A Survey on Biomedical Text Summarization with Pre-trained Language Model}.
\newblock \bibinfo{journal}{\emph{arXiv preprint arXiv:2304.08763}} (\bibinfo{year}{2023}).
\newblock


\bibitem[Xie et~al\mbox{.}(2023b)]%
        {xie2023faithful}
\bibfield{author}{\bibinfo{person}{Qianqian Xie}, \bibinfo{person}{Edward~J Schenck}, \bibinfo{person}{He~S Yang}, {et~al\mbox{.}}} \bibinfo{year}{2023}\natexlab{b}.
\newblock \showarticletitle{Faithful ai in medicine: A systematic review with large language models and beyond}.
\newblock \bibinfo{journal}{\emph{medRxiv}} (\bibinfo{year}{2023}).
\newblock


\bibitem[Xiong et~al\mbox{.}(2023)]%
        {xiong2023doctorglm}
\bibfield{author}{\bibinfo{person}{Honglin Xiong}, \bibinfo{person}{Sheng Wang}, \bibinfo{person}{Yitao Zhu}, {et~al\mbox{.}}} \bibinfo{year}{2023}\natexlab{}.
\newblock \bibinfo{title}{DoctorGLM: Fine-tuning your Chinese Doctor is not a Herculean Task}.
\newblock
\newblock
\showeprint[arxiv]{2304.01097}~[cs.CL]


\bibitem[Zhou et~al\mbox{.}(2022)]%
        {zhou2022teaching}
\bibfield{author}{\bibinfo{person}{Hattie Zhou}, \bibinfo{person}{Azade Nova}, \bibinfo{person}{Hugo Larochelle}, \bibinfo{person}{Aaron Courville}, \bibinfo{person}{Behnam Neyshabur}, {and} \bibinfo{person}{Hanie Sedghi}.} \bibinfo{year}{2022}\natexlab{}.
\newblock \bibinfo{title}{Teaching Algorithmic Reasoning via In-context Learning}.
\newblock
\newblock
\showeprint[arxiv]{2211.09066}~[cs.LG]


\bibitem[Zhu et~al\mbox{.}(2020)]%
        {zhu2020question}
\bibfield{author}{\bibinfo{person}{Ming Zhu}, \bibinfo{person}{Aman Ahuja}, \bibinfo{person}{Da-Cheng Juan}, \bibinfo{person}{Wei Wei}, {and} \bibinfo{person}{Chandan~K Reddy}.} \bibinfo{year}{2020}\natexlab{}.
\newblock \showarticletitle{Question answering with long multiple-span answers}. In \bibinfo{booktitle}{\emph{Findings of the Association for Computational Linguistics: EMNLP 2020}}. \bibinfo{pages}{3840--3849}.
\newblock


\end{thebibliography}

\end{document}